%% file: main.tex
\documentclass[10pt,twocolumn,letterpaper]{article}

\usepackage[pagenumbers]{iccv} 

\input{preamble}

\usepackage[accsupp]{axessibility}  
\definecolor{iccvblue}{rgb}{0.21,0.49,0.74}
\usepackage[pagebackref,breaklinks,colorlinks,allcolors=iccvblue]{hyperref}

\def\paperID{5} 
\def\confName{OpenSUN3D}
\def\confYear{2025}

\title{Sparse Multiview Open-Vocabulary 3D Detection}

\author{\parbox{16cm}{\centering
    {\large Olivier Moliner$^{1,2}$, Viktor Larsson$^{1}$ and Kalle Åström$^{1}$}\\
    {\normalsize
    $^1$ Centre for Mathematical Sciences, Lund University\quad
    $^2$ Sony Corporation, Lund Laboratory, Sweden}\\
    {\tt\small \{olivier.moliner, viktor.larsson, karl.astrom\}@math.lth.se}
}
}

\begin{document}
\maketitle
\input{sec/0_abstract}    
\input{sec/1_intro}
\input{sec/2_related_work}

\input{sec/3_method}

\input{sec/4_experiments}

\input{sec/5_conclusion}
{
    \small
    \bibliographystyle{ieeenat_fullname}
    \bibliography{main}
}


\end{document}

%% file: preamble.tex
\newcommand{\fst}{\cellcolor{Green!60} \bf}
\newcommand{\snd}{\cellcolor{Green!40}}
\newcommand{\thr}{\cellcolor{Green!15}}

\usepackage{overpic}
\usepackage{adjustbox}

\newcommand\T{\rule{0pt}{2.6ex}}       
\newcommand\B{\rule[-1.2ex]{0pt}{0pt}} 

\usepackage{gensymb} 
\usepackage{multirow}
\usepackage{pifont}
\newcommand{\cmark}{\ding{51}}%
\newcommand{\xmark}{\ding{55}}%

\def\OURS{SMOV3D\xspace}

\usepackage{xcolor,colortbl}
\definecolor{Red}{HTML}{FF4A30}
\definecolor{Orange}{HTML}{CF4A30}
\definecolor{Purple}{HTML}{911146}
\definecolor{Light}{HTML}{FFFF00}
\definecolor{LightBlue}{HTML}{00E5FF}
\definecolor{myblue2}{HTML}{B5EFFF}

\usepackage{tikz}
\usepackage{pgfplots}
\pgfplotsset{compat=1.17}

\newcommand{\xx}{\mathbf{x}}
\newcommand{\XX}{\mathbf{X}}

%% file: sec/0_abstract.tex
\begin{abstract}

The ability to interpret and comprehend a 3D scene is essential for many vision and robotics systems. In numerous applications, this involves 3D object detection, i.e.~identifying the location and dimensions of objects belonging to a specific category, typically represented as bounding boxes. This has traditionally been solved by training to detect a fixed set of categories, which limits its use.
In this work, we investigate open-vocabulary 3D object detection  in the challenging yet practical sparse-view setting, where only a limited number of posed RGB images are available as input.
Our approach is training-free, relying on pre-trained, off-the-shelf 2D foundation models instead of employing computationally expensive 3D feature fusion or requiring 3D-specific learning.
By lifting 2D detections and directly optimizing 3D proposals for featuremetric consistency across views, we fully leverage the extensive training data available in 2D compared to 3D. 
Through standard benchmarks, we demonstrate that this simple pipeline establishes a powerful baseline, performing competitively with state-of-the-art techniques in densely sampled scenarios while significantly outperforming them in the sparse-view setting.

\end{abstract}

%% file: sec/1_intro.tex
\section{Introduction}
\label{sec:intro}
The ability to parse and understand a 3D scene is a prerequisite for many vision or robotic systems.
In many applications, this takes the form of 3D object detection, i.e.~determining the location and dimension of all objects of a particular category, e.g.~as a bounding box.
Object detection is a classical problem in computer vision, and is traditionally solved by selecting a discrete set of object categories, which the method is trained to detect.
Having a fixed set of labels limits the applicability of the methods and prevents them from generalizing to new problem domains not represented in the label-set without expensive re-training.

Recent advances in 2D visual-language models have allowed for so-called \textit{open-vocabulary} object detection, where the system can be queried with arbitrary labels.
The expressive power coming from incorporating the language model also allows for more complex queries, e.g.~a lengthier description of an object beyond single labels or even via object affordances (\textit{Where can I sit?}).
While this development originally happened in 2D, several methods have been proposed for open-vocabulary 3D object detection using these rich feature representations.

\begin{figure}[t]
    \centering
    \includegraphics[width=\columnwidth]{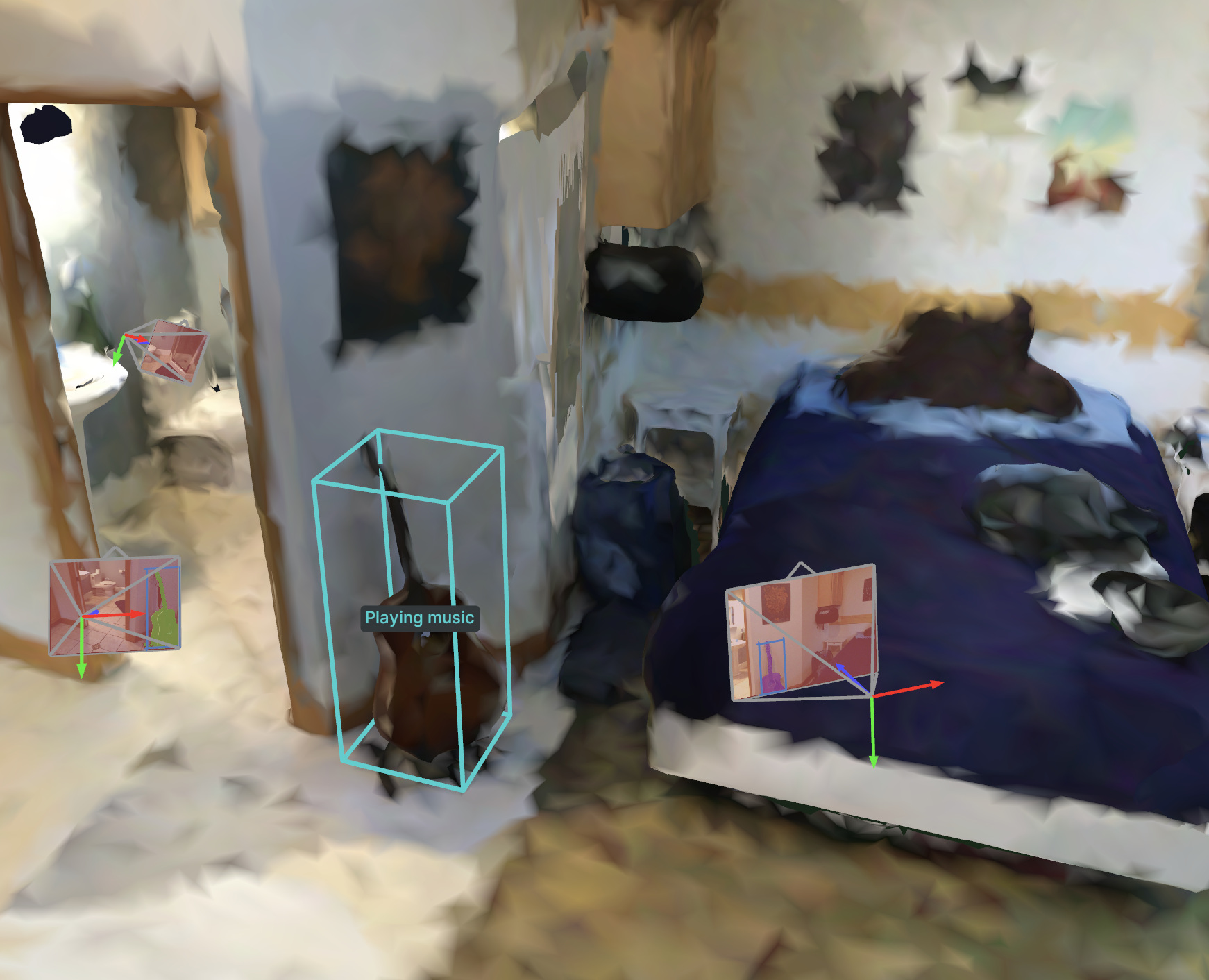}    
    \caption{\textbf{Open-vocabulary 3D Object Detection.} Our method takes as input a sparse collection of posed RGB images together with a query text prompt. The outputs are 3D bounding boxes corresponding to the prompt. In the figure we include the ground-truth mesh for visualization purposes only. }
    \label{fig:teaser}
\end{figure}

\input{figures/overview_fig}
Existing methods either rely on dense 3D geometry obtained by scanning and reconstructing the scene offline, or perform monocular detection in RGB-D images.
To detect new or moved objects, the scene needs to be rescanned, hence many applications requiring continuous monitoring of a scene are not feasible in this setting.
Most methods use
trained 3D proposal networks to localize objects in the 3D data. 
They leverage vision-language models either when training the 3D backbone or at inference time by back-projecting language-augmented visual features and matching open-vocabulary 2D detections to class-agnostic 3D masks.
However, current 3D datasets are orders of magnitude smaller than the large-scale image datasets used to train 2D foundation models. 
As a result, the strong generalization capabilities of the 2D models may be lost during training. 

In this work, we investigate how effectively 2D foundation models can be leveraged to tackle the task of open-vocabulary 3D object detection from only a sparse set of RGB images, without any 3D-specific training, and introduce a straightforward, training-free baseline.
Our approach works by lifting detections from off-the-shelf 2D open-vocabulary and segmentation models into 3D via monocular depth, followed by a multi-view refinement that optimizes for photometric and semantic consistency. 
Our experiments show that this simple approach is surprisingly powerful, producing comparable results to state-of-the-art methods in densely-sampled scenarios while
establishing a strong baseline for the sparse-view setting.

%% file: figures/overview_fig.tex
\begin{figure*}[t!]
    \centering

     \begin{overpic}[width=\textwidth]{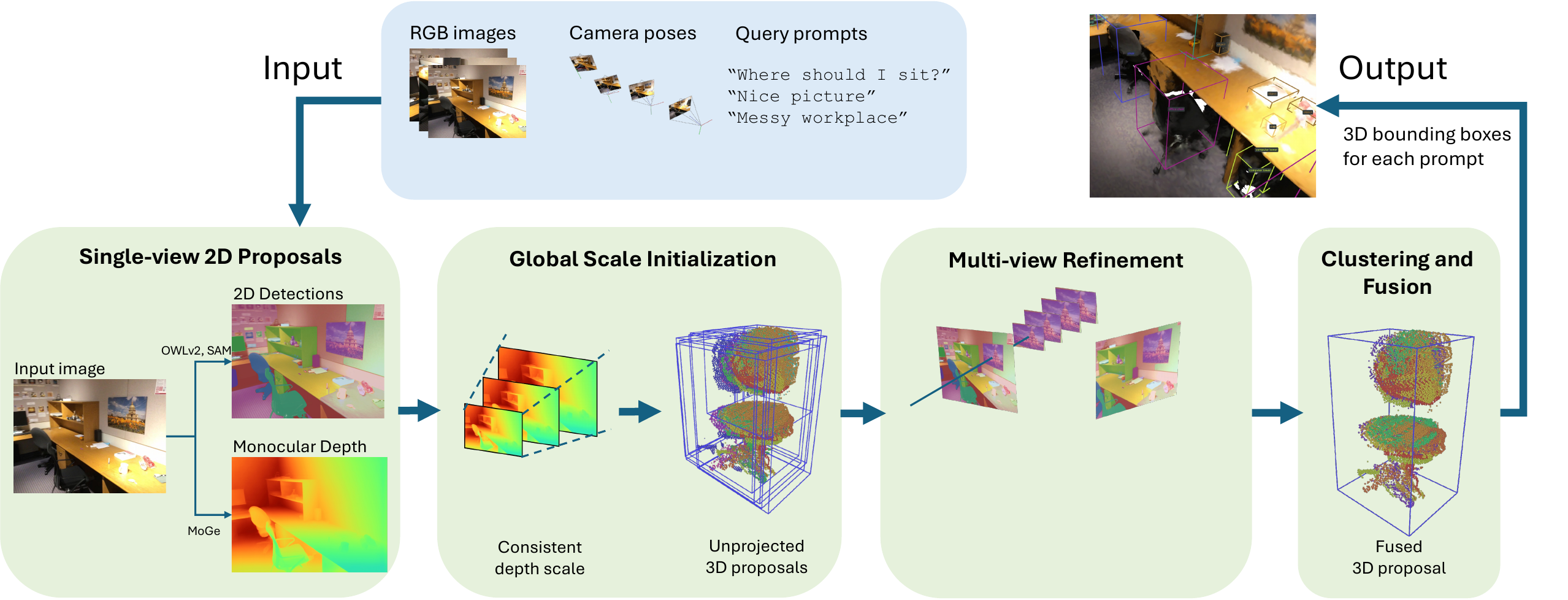}
    \put(9,-2.5){Section~\ref{sec:mono_proposals}}
    \put(35,-2.5){Section~\ref{sec:scale_finding}}
    \put(30.5, 5){\small $D_i = \alpha d_i$}
    \put(63,-2.5){Section~\ref{sec:per_mask_refinement}}
    \put(85,-2.5){Section~\ref{sec:proposal_fusion}}
    \put(25, 27.5){\small $\mathcal{I}_1,\dots,\mathcal{I}_N$}
    \put(35, 27.5){\small $\{(R_i, t_i, K_i)\}_{i=1}^N$}
    \put(57,8){\resizebox{!}{0.25cm}{$\displaystyle\min_{\alpha,\beta} \sum_i \sum_k \left\| \mathcal{I}_i\left[ \xx_{ik}^{proj}(\alpha,\beta)\right] - \mathcal{I}_0\left[ \xx_k \right] \right\|_1$}}
    \put(58.6,4.6){\resizebox{!}{0.27cm}{$\displaystyle + \sum_i \sum_k \left\| \mathcal{F}_i\left[ \xx_{ik}^{proj}(\alpha,\beta)\right] - \mathcal{F}_0\left[ \xx_k \right] \right\|^2 $}}
    \end{overpic}\\~\\
    
    \caption{\textbf{Overview of SMOV3D.} Our method takes as input a sparse collection of posed RGB images together with a collection of text query prompts. The pipeline then conists of three steps. \textbf{i) Monocular 2D Proposals} For each prompt and image we perform 2D detection yielding a set of masks. These are then lifted to 3D using monocular depth. \textbf{ii) Multi-view Refinement} Lifted 3D point clouds are refined by optimizing a multi-view featuremetric loss that combines both photometric and CLIP consistency. \textbf{iii) 3D Clustering and Fusion.} The optimized 3D point clouds are aggregated in 3D and greedily fused using a simple heuristic. The output is a collection of 3D bounding boxes. For visualization we show them overlayed on the ground-truth mesh.}
    \label{fig:overview}
\end{figure*}
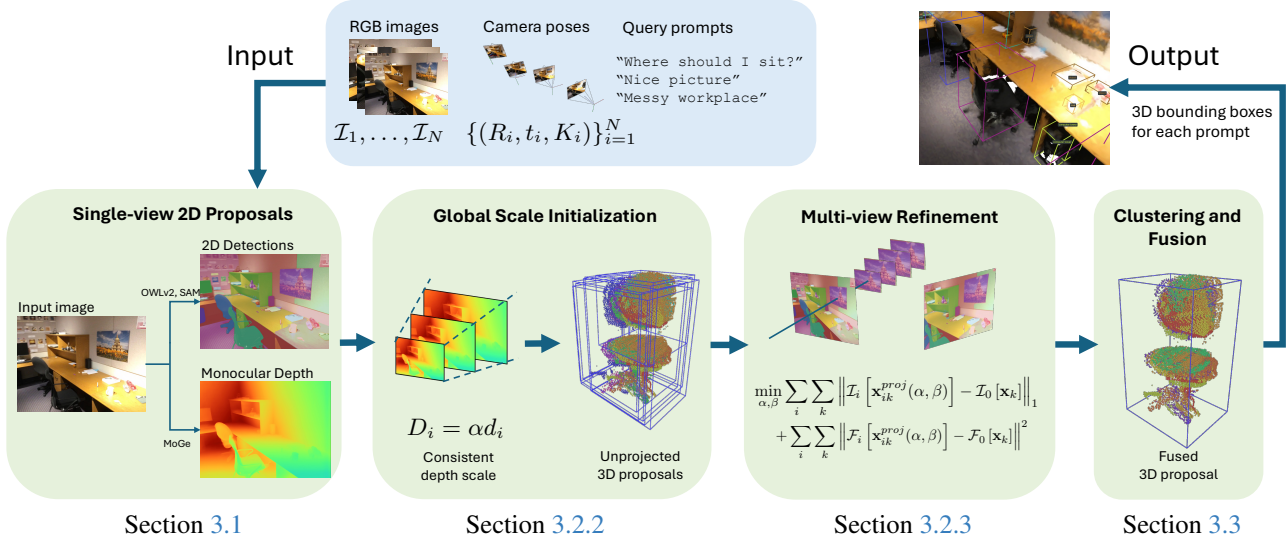

%% file: sec/2_related_work.tex
\section{Related Work}
\label{sec:related}

{\flushleft \bf 3D object detection.}
3D object detection aims at predicting three-dimensional bounding boxes and object classes from 3D or 2D input data.
3D point clouds are appealing for 3D object detection as they provide accurate geometric information, and point cloud-based methods leveraging deep Hough voting \cite{Qi2019DeepHV,Cheng2021BacktracingRP,Zhang2020H3DNet3O,Wang2022RBGNetRG} or Transformers \cite{Misra2021AnET,Liu2021GroupFree3O} have shown remarkable performance in both indoor and outdoor settings.
However, these methods require depth sensors or dense image sequences for data acquisition, which may not be practical due to cost or power consumption constraints. 

Early methods for 3D object detection from RGB images extend conventional 2D detectors to lift monocular detections to 3D \cite{Simonelli2019DisentanglingM3, Shi2021GeometrybasedDD, Qin2018MonoGRNetAG,Lu2021GeometryUP}, but their performance suffers from the lack of explicit depth information.
Recently, methods leveraging multiple views to better capture scene geometry have gained increasing attention.
Extending DETR \cite{Carion2020EndtoEndOD} to the 3D domain, Transformer-based methods \cite{Wang2021DETR3D3O, Liu2022PETRPE, Tseng2022CrossDTRCA, Xie2023PixelAlignedRQ} predict bounding boxes by attending to multi-view features.
Feature volume-based methods \cite{Rukhovich2021ImVoxelNetIT,Tu2023ImGeoNetIG} perform 3D object detection in a voxel-based 3D feature volume \cite{Murez2020AtlasE3},
while recent work leverages Neural Radiance Fields (NeRF) \cite{mildenhall2020nerf} to model geometry implicitly in the feature volume \cite{xu2023nerf}.

Despite their proven performance, current state-of-the-art 3D object detectors, whether 3D or 2D-based, are closed-set methods trained to detect a limited number of predefined object categories. Extending these models to new domains requires collecting and annotating new data and retraining the models, which is both costly and time-consuming.

{\flushleft \bf Open-vocabulary 2D and 3D object detection.}
Open-vocabulary object detection is an emerging field in computer vision that aims to localize and identify arbitrary, previously unseen objects.
The ideas of going from closed to open vocabulary tasks emerged with the advent of better foundation models, for example GPT for text, \cite{NEURIPS2020_1457c0d6}, CLIP for image-text \cite{clip}, and Lidar-CLIP for image-text-lidar \cite{hess2024lidarclip}. Such models have been shown to be important for a large variety of few- or single-shot learning and open vocabulary tasks. For example CLIP opened up for open-ended queries primarily for classification and action recognition, but later open-vocabulary 2D Object detection was explored and developed, e.g.~in \cite{zhou2022extract,rao2022denseclip,Chen_2023_open,ding2023maskclip}. 

Open-vocabulary 3D object detection is still in its infancy.
Existing open-vocabulary 3D detection models usually operate on point-cloud or RGB-D data.
OV-3DETIC \cite{lu2022open} expands a 3D object detector's vocabulary using ImageNet1K \cite{Deng2009ImageNetAL} and uses contrastive learning to transfer knowledge between image and point cloud modalities.
OV-3DET \cite{lu2023open} generates pseudo-annotations using a pre-trained 2D open-vocabulary detector \cite{Zhou2022DetectingTC} to train a 3D detector to localize objects.
Object2Scene \cite{zhu2023object2scene} proposes a point-cloud object detector (L3Det) trained on a 3D dataset augmented by inserting 3D objects and corresponding text descriptions. It leverages cross-domain contrastive learning to mitigate the domain gap between scene and inserted objects.
FM-OV3D \cite{zhang2023fmov3d} blends knowledge from multiple pre-trained foundation models to improve the open-vocabulary localization and recognition abilities of its 3D detection model.
OpenIns3D \cite{huang2024openins3d} casts the problem as an extension of open-vocabulary semantic segmentation. Its "Snap" module generates synthetic images from point clouds and uses 2D vision-language models to detect objects in 2D based on text prompts, while its "Lookup" module matches the 2D detections to class-agnosic 3D point clounds predicted by the "Mask" module.
ImOV3D \cite{yang2024imov3d} addresses the scarcity of annotated 3D datasets by generating pseudo 3D point clouds and annotations from 2D datasets to train an open-vocabulary point cloud detector.

The reliance on point cloud data, whether obtained by depth sensors or dense image sequences, limits the applicability of open-vocabulary 3D detection, and the need for trained 3D proposal networks raises the question of the ability of existing methods to generalize to new domains.
In this work, we leverage pre-trained foundation models to introduce a straightforward, training-free method for open-vocabulary 3D object detection using only sparse, multi-view RGB images as input.

%% file: sec/3_method.tex
\section{Method}
\label{sec:method}

 We now present our method for open-vocabulary 3D object detection which we call \textbf{S}parse \textbf{M}ulti-view \textbf{O}pen-\textbf{V}ocabulary \textbf{3}D \textbf{D}etection (\OURS).
Our method takes as input a collection of RGB images $\{\mathcal{I}_i\}_{i=1}^m$ together with poses and intrinsics $\{(R_i,t_i,K_i)\}_{i=1}^m$, as well as query text prompts.
For each of the prompts, the method then consists of three steps:
\begin{itemize}
    \item  For each image we generate a collection 2D proposals which are lifted to a camera-centric point cloud using monocular depth estimation. (Sec.~\ref{sec:mono_proposals})
    \item Each proposal is then refined by considering a multi-view featuremetric consistency. (Sec~\ref{sec:multiview_refinement})
    \item The proposals are then robustly clustered in 3D and fused to generate the final result. (Sec.~\ref{sec:proposal_fusion})
\end{itemize}
The method returns a collection of 3D bounding boxes. For each prompt, multiple bounding boxes can be returned if there are multiple instances present in the scene.
Figure~\ref{fig:overview} shows an overview of our method and in the next sections we detail each of the three steps.

\subsection{Single-view Proposal Generation} \label{sec:mono_proposals}
In a first step, we generate initial 3D object proposals for each 2D view.

In each image, we generate a collection of 2D object bounding boxes using a state-of-the-art 2D open-vocabulary detector (OWLv2~\cite{minderer2023scaling}) queried with each prompt.
These bounding boxes are used as input to an image segmentation model (Segment Anything~\cite{Kirillov_2023_ICCV}) to produce accurate 2D masks.
Finally, we lift each 2D mask to 3D using an affine-invariant monocular depth estimator (MoGe~\cite{wang2024moge}).

As monocular depth estimators tend to oversmooth edges, we first filter out depth values with high gradients (i.e.~where normals are close to orthogonal with the viewing direction).
Due to image noise, or to the presence of several similar objects in front of each other, the 2D masks generated by SAM sometimes contain parts of different objects.
To separate erroneously merged objects and remove background points, we cluster the 3D proposals with DBSCAN \cite{ester1996density}. We treat large clusters as new proposals, and remove small clusters and outlier points.

\subsection{Multi-view Proposal Refinement} \label{sec:multiview_refinement}
The monocular depth estimator predicts relative depth maps that are invariant to affine (scale and shift) transformations.
Moreover, the depth maps are often only locally consistent, thus even if the optimal global shift and scale parameters are estimated, the depths of individual objects may be inaccurate.
To retrieve accurate 3D positions and sizes for the 3D proposals, we first find an initial global scale factor for each input image using multi-view semantic consistency. 
We then proceed to refine each 3D mask by optimizing individual scale and shift parameters.

\subsubsection{Multi-view Consistency of 3D Proposals}
To accurately estimate the 3D positions of the detected objects we require a multi-view consistency loss, measuring how much the backprojected proposals from each image are supported by the other images.

Let $\{\xx_k\}_{k=1}^N$ be the pixels belonging to a proposal in the image $\mathcal{I}_0$. 
These are lifted to 3D in the camera coordinate system by backprojecting using the depth as
\begin{equation}
    \XX^{cam}_k = (\alpha d_k + \beta) K_0^{-1} \xx_k ,
\end{equation}
where $d_k$ is the original mono-depth depth value, and $\alpha,\beta$ are the shift-scale parameters which we aim to  recover.
Using the known camera poses and intrinsics, the lifted 3D point can be projected into the $i$th view $\mathcal{I}_i$ as
\begin{equation}
    \xx_{ik}^{proj} = \Pi\left(K_i(R_i R_0^T (\XX^{cam}_k - t_0) + t_i)\right),
\end{equation}
where $\Pi : \mathbb{R}^3 \to \mathbb{R}^2$ is the pinhole projection function.
Note that $\xx_{ik}^{proj}$ is now a function of scale $\alpha$ and shift $\beta$.

To measure consistency across images we introduce a photo-consistency loss as
\begin{equation}
    \mathcal{L}_{rgb}(\alpha,\beta) = \sum_i \sum_k \left\| \mathcal{I}_i\left[ \xx_{ik}^{proj}(\alpha,\beta)\right] - \mathcal{I}_0\left[ \xx_k \right] \right\|_1,
\end{equation}
where $\mathcal{I}_i[\xx] \in \mathbb{R}^3$  denotes the image $\mathcal{I}_i$ (bi-linearly) interpolated at the pixel position $\xx$.
The loss is normalized over the reprojected points visible in $\mathcal{I}_i$.
To improve robustness to viewpoint changes and other effects causing photometric inconsistencies across images we also include a CLIP-based consistency term in the optimization.
For each image we compute a dense CLIP feature map,
\begin{equation}
    \mathcal{F}_i = \text{CLIP}(\mathcal{I}_i)
\end{equation}
and then define the reprojected CLIP loss as 
\begin{equation}
    \mathcal{L}_{sim}(\alpha,\beta) = \sum_i \sum_k \left\| \mathcal{F}_i\left[ \xx_{ik}^{proj}(\alpha,\beta)\right] - \mathcal{F}_0\left[ \xx_k \right] \right\|^2 .
\end{equation}
This term will ensure alignment to similar semantic image content in cases where the photometric loss is insufficient.

For each 3D proposal, the full consistency measure is then a weighted combination of these two,
\begin{equation} \label{eq:full_loss}
    \mathcal{L}(\alpha, \beta) = \mathcal{L}_{rgb}(\alpha, \beta) + \lambda \mathcal{L}_{sim}(\alpha, \beta) ,
\end{equation}
where $\lambda$ is the trade-off hyperparameter.

\subsubsection{Global Scale Initialization} \label{sec:scale_finding}
In the first step of the shift-and-scale estimation we estimate a global scaling parameter $\alpha_{global}$ which is used to initialize the scale $\alpha$ for each proposal in the second step.

To do this we jointly consider all object proposals and select $\alpha_{global}$ by sampling.
Let $\alpha_1, \dots, \alpha_m$ be $m$ scales uniformly sampled in the range $[\alpha_{min}, \alpha_{max}]$, and let $\mathcal{L}_k$ denote the loss \eqref{eq:full_loss} for the $k$th proposal.
We then take
\begin{equation}
    \alpha_{global} = \arg\min \left\{ \sum_k \mathcal{L}_k (\alpha, 0) ~\Bigg|~ \alpha \in \{\alpha_1,\dots,\alpha_k \} \right\} .
\end{equation}
As we are only interested in a coarse scale estimate,  we only consider at most $M$ pixels across all proposals to speed up the process.
The motivation for sampling within the detected 2D masks instead of anywhere in the image is to avoid sampling in background areas, e.g. walls, ceiling, sky, which are not as helpful for estimating multi-view consistency.

\subsubsection{Per-mask Refinement} \label{sec:per_mask_refinement}
The previous step yields an initial estimate of the scale parameter $\alpha$, but to allow for errors in the depth map, we now optimize an independent $\alpha$ and $\beta$ for each proposal.
This allows us to handle cases where the depth map is not globally consistent across different scene elements, which is supported by our ablations (Section~\ref{sec:ablations}).
Each proposal is then refined by minimizing 
\begin{equation}
        \min_{\alpha,\beta}~~\mathcal{L}(\alpha,\beta),
\end{equation}
initialized with $\alpha_{global}$ and zero. 
The optimization is performed using gradient descent.
To speed up the optimization, we randomly sample a subset of the mask pixels at each iteration.
Note that as each proposal is optimized independently, this can be done in parallel.

\subsection{Clustering and Fusion} \label{sec:proposal_fusion}

For each image and prompt, we now have a collection of 3D proposals (point clouds).
The last step of the pipeline is now to combine these into the final output bounding boxes.

For this, we use a simple sequential clustering approach.
First, we compute axis-aligned 3D bounding boxes for all proposals.
These are then greedily merged based on their intersection-over-union (IoU).
 Finally, for each merged cluster we compute a bounding box for the union of the point-clouds.
\vspace{-2pt}

%% file: sec/4_experiments.tex
\input{tables/tab_scannet_10}
\input{tables/tab_scannet_20}
\section{Experiments}
\label{sec:exp}


\subsection{Experimental setup}
{\flushleft \bf Datasets.}  
The ScanNet dataset \cite{dai2017scannet} comprises 1,201 training and 312 validation scenes.
We evaluate our method on the ScanNet10 and ScanNet20 categories defines by Lu \etal \cite{lu2022open,lu2023open}.
To demonstrate the open-vocabulary capabilities of our method, we show results on the ScanNet200 benchmark \cite{rozenberszki2022language}, in which the 200 most represented object classes of ScanNet are split into 3 subsets based on the frequency of the number of labeled surface points in the training set: \textit{head} (66 classes), \textit{common} (68 classes) and \textit{tail} (66 classes). 
We also experiment with Replica \cite{straub2019replica}, a dataset of photo-realistic 3D indoor scenes reconstructed from RGB-D scans, which contains 48 object classes.
To demonstrate the ability of our method to handle arbitrary queries, we present qualitative results on data from the OpenSUN3D challenge \cite{OpenSUN3D}, featuring ARKitScenes scans \cite{baruch2021arkitscenes} with long-tail prompts.

{\flushleft \bf Evaluation protocol.}
We follow the class splits of prior works \cite{rozenberszki2022language,lu2023open,huang2024openins3d}, without using "seen" classes.
We compute axis-aligned bounding boxes from ground-truth segmentations, following \cite{huang2024openins3d}.
We report the performance on the validation sets using the mean Average Precision at an IoU threshold of $0.25$, denoted $mAP_{25}$.

{\flushleft \bf Implementation details.} 
For each scene, we sample 2D views such that each frame has a relative translation greater than $0.5m$ or a relative rotation angle greater than $15\degree$ from any previously sampled frame.
Unless otherwise stated, we then sample at most 32 random views to cover a scene. We present results averaged over three random seeds for sampling the views. 
We use OWLv2 \cite{minderer2023scaling} as open-vocabulary 2D object detector, and SAM2 \cite{Kirillov_2023_ICCV, ravi2024sam2} to produce the 2D masks.
For monocular depth we use MoGe~\cite{wang2024moge}.
To extract dense CLIP feature maps, we use the MaskCLIP \cite{zhou2022extract} reparametrization trick with CLIP ViT-L/14.
During depth refinement, we use the AdamW optimizer \cite{loshchilov2018decoupled} with a learning rate of 0.005 and sample 100 points randomly at each iteration.
The weight $\lambda$ is set to $1.0$.
We tuned our method's hyperparameters on a subset of ScanNet's training set.

{\flushleft \bf Baselines.} 
Since no previous work has tackled open-vocabulary 3D object detection from sparse multi-view 2D images, we compare our method to state-of-the-art point cloud-based methods: OpenIns3D \cite{huang2024openins3d}, FM-OV3D \cite{zhang2023fmov3d}, Object2Scene~\cite{zhu2023object2scene}, OV-3DETIC~\cite{lu2022open} and OV-3DET~\cite{lu2023open}.
For comparison, we also evaluate our method using the depth maps provided by the datasets. In this setting, we use the same view sampling as for our RGB-based method, and do not perform depth refinement. 

\subsection{Quantitative Comparison}

{\flushleft \bf ScanNet.}
We first evaluate our method on the ScanNet10 and ScanNet20 benchmarks.
As can be seen in \cref{table:scannet_10} and \cref{table:scannet_20}, when using ground-truth depth maps our method is comparable to OpenIns3D and surpasses all other point cloud-based methods.
In this setting, our method is very simple, as it only involves backprojecting 2D mask proposals.
Conversely, all other point cloud-based methods use a 3D box or mask proposal network trained on ScanNet.

Somewhat surprinsingly,
when using only sparse 2D views, our method still surpasses most existing point cloud-based methods.
Some are monocular methods \cite{zhu2023object2scene,lu2022open,lu2023open} that rely on pseudo-3D annotations constrained to the view frustum during training. As ScanNet images have a narrow field of view, many objects are truncated and may have varying shapes during training, which may lead to lower performance.
As MoGe was not trained on ScanNet, our method is the only truly zero-shot method in this benchmark.

{\flushleft \bf Replica.} 
\input{tables/tab_replica}
We also evaluate our method on the Replica dataset, 
using the same hyperparameters as for ScanNet, 
to confirm its ability to generalize.
Although there are no previous results for open-vocabulary 3D object detection on Replica, OpenIns3D is easily adaptable to this task, by converting generated masks into axis-aligned bounding boxes. We also evaluated OpenIns3D with RGB-D data, using the same settings used for open-vocabulary instance segmentation on Replica in the original implementation. 
In this setting, OpenIns3D takes as input the point cloud and 200 RGB-D images per scene.

As shown in \cref{table:replica}, the performance of our method on Replica is comparable to its performance on ScanNet, demonstrating its ability to generalize.
On the other hand, OpenIns3D's performance dropped. This decrease is compatible with the decrease observed on the segmentation task in the original paper, and might be due to the mask proposal model's struggle to generalize to Replica.

\input{figures/qualitative}

{\flushleft \bf Long-tail 3D Object Detection.} To study our method's zero-shot generalization ability, we evaluate it on the ScanNet200 benchmark. 
During evaluation, we use the whole ScanNet200 vocabulary (except, as is usual, \textit{'wall'} and \textit{'floor'}), and present results for the \textit{Head}, \textit{Common} and \textit{Tail} category splits.
We compare with Object2Scene \cite{zhu2023object2scene}, which used the \textit{Head} as seen classes, and OpenIns3D.
Following the authors' advice, we only present results for OpenIns3D with RGB-D data.
The results can be seen in \cref{table:scannet_200}.
\OURS achieves strong performance on tail categories, even surpassing its performance on common categories. 
While counter-intuitive, this can be explained by our training-free approach.
Unlike methods trained or fine-tuned on 3D datasets, \OURS relies solely on the generalization of the 2D foundation models and does not inherit the dataset's label frequency biases.
\input{tables/tab_scannet_200}
\input{figures/fig_ablation_nviews}

\subsection{Qualitative Results} \label{sec:qualitative}
In \cref{fig:teaser}, we show an example of 3D object detection with a free-text prompt (\textit{"Playing music"}) in a ScanNet scene. The figure also shows posed cameras and 2D mask proposals.
In \cref{fig:qualitative}, we visualize 3D bounding boxes predicted by our method when prompted with the ScanNet10 categories.

\Cref{fig:arkitscenes} presents qualitative results using scenes and queries from the OpenSUN3D challenge \cite{OpenSUN3D}, which features ARKitScenes \cite{baruch2021arkitscenes} scans paired with long-tail prompts.
It demonstrates the ability of our method to perform zero-shot detection from free-form text queries in realistic scenes.

\input{figures/arkitscenes}
\input{figures/replica}

\subsection{Using Few Views} \label{sec:few_views}
Current open-vocabulary 3D detection benchmarks rely either on 3D scenes  reconstructed offline, or on monocular RGB-D images following a scanning trajectory.
We argue that these approaches are
ill-suited for many practical use cases, as detecting new or moved objects would require a complete rescan of the scene.
A more realistic setting for many applications such as facility management or retail is to consider systems with few fixed, pre-calibrated cameras, delivering predictions at regular intervals.

To address this gap, we 
emulate a fixed-camera installation in the Replica dataset by placing 4 cameras at room corners near the ceiling and recursively placing a camera between each pair to obtain 8 and 16 views.
We compare to OpenIns3D using the same views, and to OpenIns3D with Snap \& Lookup.
Note that 
OpenIns3D still uses the full point cloud for mask proposal, including points that are not visible in any of the views.

\OURS takes 15.6, 26.7 and 50.5 seconds to perform full-scene detection on an RTX 4090 with 4, 8 and 16 RGB views respectively.
As shown in \cref{tab:ablation_nviews}, with as few as 4 RGB images, our method significantly outperforms the point-cloud-based OpenIns3D.
We also present qualitative results in \cref{fig:replica}, showing camera placement and predicted bounding boxes.

This experiment shows that this simple approach is well-suited for applications requiring continuous monitoring under practical constraints.

\subsection{Ablation Study} \label{sec:ablations}

In \cref{table:ablation_refinement}, we analyze key components of our method on ScanNet10.
With global scale initialization only, i.e.~estimating one scale parameter for each view, our method, though simple, already achieves results comparable to point cloud-based methods.
Refining the depth maps for each object mask separately using any combination of $\mathcal{L}_{rgb}$ and $\mathcal{L}_{sim}$ provides an improvement. We believe this is due to the depth maps only being locally consistent.

We also implemented a depth refinement method minimizing a multi-view depth consistency loss $\mathcal{L}_{depth}$, defined as the $L_1$ loss between the estimated depth of proposal points in one view and the depths of their reprojections in other views.
This leads to degraded results, further showing that the initial monocular depth maps are geometrically inconsistent.
The best performance is obtained by refining the depth maps for each object mask separately using both the photometric loss $\mathcal{L}_{rgb}$ and the CLIP similarity loss $\mathcal{L}_{sim}$, yielding a 12\% relative improvement in the mAP$_{25}$ metric.
\input{tables/tab_ablation_refinement}

%% file: tables/tab_scannet_10.tex
\begin{table*}[t]
\small
\centering
\resizebox{1.0\textwidth}{!}{
\begin{tabular}{lcc r| rrrrrrrrrr}
\toprule[1pt]
Method & GT Depth & 3D proposal      & Mean & toilet & bed & chair & sofa & dresser & table & cabinet & bookshelf & pillow & sink\T\B\\
\midrule[.6pt]
OV-3DETIC~\cite{lu2022open} & \textcolor{Red}{\textbf{\cmark}} &  3DETR$^\dagger$   & 12.7 & 49.0   & 2.6  & 7.3   & 18.6 & 2.8     & 14.3  & 2.4     & 4.5       & 3.9    & 21.1 \\ 
Object2Scene~\cite{zhu2023object2scene} & \textcolor{Red}{\textbf{\cmark}} & L3DET$^\dagger$ & 24.6 & 56.3   & \thr 36.2 & 16.1  & 23.0 & 8.1     &  \thr 23.1  & \snd 14.7    & 17.3      & \thr 23.4   & 27.9 \\ 
FM-OV3D \cite{zhang2023fmov3d} & \textcolor{Red}{\textbf{\cmark}} & 3DETR$^\dagger$ & 21.5 & 55.0   & \snd 38.8 & \snd 19.2  & \thr 41.9 & \snd 23.8    & 3.5   & 0.4     & 6.0       & 17.4   & 8.8  \\ 
OpenIns3D \cite{huang2024openins3d} & \textcolor{Red}{\textbf{\cmark}} & Mask3D$^\dagger$ & \fst 43.7 & \snd 79.5   & \fst 70.5 & \fst 76.9  & 15.8 & 0.0     & \fst 53.1  & \fst 40.1    & \fst 41.2      & 7.1    & \snd 53.1\\
\OURS (Ours)  \tiny{RGB-D} & \textcolor{Red}{\textbf{\cmark}} & - & \snd 42.2  & \fst 83.5 & 29.9 & \snd 29.8 & \fst 73.5   & \fst 25.8 & \snd 27.8    & \thr 4.1     & \snd 28.4  & \fst 61.1 & \fst 58.6\B\\ 
\midrule[.2pt]
\OURS (Ours) & \textcolor{ForestGreen}{\xmark} & - & \thr 28.9  & \thr 61.8 & 25.9 & 12.2 & \snd 61.0   & \thr 16.2 & 20.0    & 1.0     & \thr 23.3  & \snd 36.1 & \thr 31.8\B\T\\ 
\bottomrule[1pt]
\end{tabular}
}
\caption{\label{table:scannet_10}\textbf{Open-vocabulary Object Detection on ScanNet10.} We compare our method to point cloud-based methods (mAP$_{25}$, \%). 
}
\end{table*}

%% file: tables/tab_scannet_20.tex
\begin{table*}[t]
\small
\centering
\resizebox{\textwidth}{!}{
\begin{tabular}{lcr|rrrrrrrrrrrrrrrrrrrr}
\toprule[1pt]
\textbf{Methods}   &     \textbf{GT depth}      & \textbf{Mean} & \rotatebox{45}{toilet}	&\rotatebox{45}{bed}	&\rotatebox{45}{chair}	&\rotatebox{45}{sofa}	&\rotatebox{45}{dresser}	&\rotatebox{45}{table}	&\rotatebox{45}{cabinet}	&\rotatebox{45}{bookshelf}	&\rotatebox{45}{pillow}	&\rotatebox{45}{sink}	&\rotatebox{45}{bathtub}	&\rotatebox{45}{refrigerator}	&\rotatebox{45}{desk}	&\rotatebox{45}{night stand}	&\rotatebox{45}{counter}	&\rotatebox{45}{door}	&\rotatebox{45}{curtain}	&\rotatebox{45}{box}	&\rotatebox{45}{lamp}	&\rotatebox{45}{bag}\T\B\\
\midrule[.6pt]
        CLIP-3D  \cite{clip}& \textcolor{Red}{\textbf{\cmark}}& 12.7 & 44.8 & 23.8 & 17.5 & 12.6 & 4.9 & 13.2 & 1.9 & 4.0 & 11.4 & 17.6 & 32.2 & 14.9 & 11.4 & 2.4 & 0.5 & \thr 14.5 & 8.6 & \thr 7.5 & 5.1 & 4.7 \T\\ 
        OV-3DET \cite{lu2023open}& \textcolor{Red}{\textbf{\cmark}} & 18.0 & \thr 57.3 & \snd 42.3 & \thr 27.1 & \thr 31.5 & 8.2 & 14.2 & \thr 3.0 & 5.6 & \thr 23.0 & \thr 31.6 & \snd 56.3 & 11.0 & 19.7 & 0.8 & 0.3 & 9.6 & \thr 10.5 & 3.8 & 2.1 & 2.7 \\
        OpenIns3D \cite{huang2024openins3d}& \textcolor{Red}{\textbf{\cmark}} & \fst 37.1 & \snd 79.5 & \fst 70.5 & \fst 76.9 & 15.8 & 0.0 & \fst 53.1 & \fst 40.1 & \fst 41.2 & 7.1 & \snd 53.1 & 14.3 & \fst 32.1 & \snd 29.1 & \thr 4.8 & \fst 55.6 & \fst 40.4 & \fst 41.1 & 2.6 & \fst 48.0 & \thr 6.2 \B\\
        \OURS (Ours)  \small{RGB-D}& \textcolor{Red}{\textbf{\cmark}} & \snd 36.2  & \fst 85.0 & 28.7 & \snd 29.9 & \fst 74.2   & \fst 28.9 & \snd 20.1    & \snd 3.8     & \snd 26.3  & \fst 60.9 & \fst 60.4   & \fst 65.9        & \snd 27.5 & \fst 32.8      & \fst 53.4   & \snd 14.4 & \snd 22.5   & \snd 15.0 & \fst 14.6 & \snd 13.7 & \fst 45.8\B\\
                \midrule
        \OURS (Ours) & \textcolor{ForestGreen}{\xmark} & \thr 21.7  & 54.5 & \thr 29.0 & 12.0 & \snd 59.9   & \snd 19.1 & \thr 16.3    & 1.0     & \thr 21.0  & \snd 35.7 & 25.7   & \thr 36.3        & \thr 15.1 & \thr 22.4      & \snd 33.7    & \thr 7.1 & 2.6    & 7.8 & \snd 10.7 & \thr 5.5 & \snd 19.4\B\T\\ 
        \bottomrule
      \end{tabular}
}
  \caption{\label{table:scannet_20}\textbf{Open-vocabulary 3D Object Detection on ScanNet20.} (mAP$_{25}$, \%).}
  \label{tab:SOTASCANNET}
  \vspace{-6pt}
\end{table*}

%% file: tables/tab_replica.tex
\begin{table}[t]
\small
\centering
\begin{tabular}[t]{l c r}
\toprule[1pt]
Method & GT depth             & mAP$_{25}$\T\B\\
\midrule[.5pt]
OpenIns3D \cite{huang2024openins3d} \tiny{point-cloud + Snap} & \textcolor{Red}{\textbf{\cmark}}      &   21.1\T\\ 
OpenIns3D \cite{huang2024openins3d} \tiny{point-cloud + RGB-D} & \textcolor{Red}{\textbf{\cmark}}      &  32.9\\ 
\OURS (Ours)  \tiny{RGB-D} &  \textcolor{Red}{\textbf{\cmark}}  & \textbf{38.9}\B\\ 
\midrule[.2pt]
\OURS (Ours) & \textcolor{ForestGreen}{\xmark}       &  29.3\B\T\\ 
\bottomrule[1pt]
\end{tabular}
\caption{\label{table:replica}\textbf{Open-vocabulary Object Detection on Replica.} 
The performance of our method on Replica is comparable to its performance on ScanNet, demonstrating its ability to generalize. Our approach has significantly higher mAP than OpenIns3D using known depth either from RGB-D or from ground truth point-clouds. Even without ground truth depth, it is competitive. 
}
\end{table}

%% file: figures/qualitative.tex
\begin{figure*}[t!]
\scriptsize
\centering
\resizebox{0.85\linewidth}{!}{
 \begin{tabular}{@{}c@{\hspace{1mm}}c@{\hspace{1mm}}c@{\hspace{1mm}}c@{\hspace{1mm}}c@{}}
   \rotatebox[origin=c]{90}{Ground truth} &
   \raisebox{-0.5\height}{\includegraphics[width=0.25\textwidth]{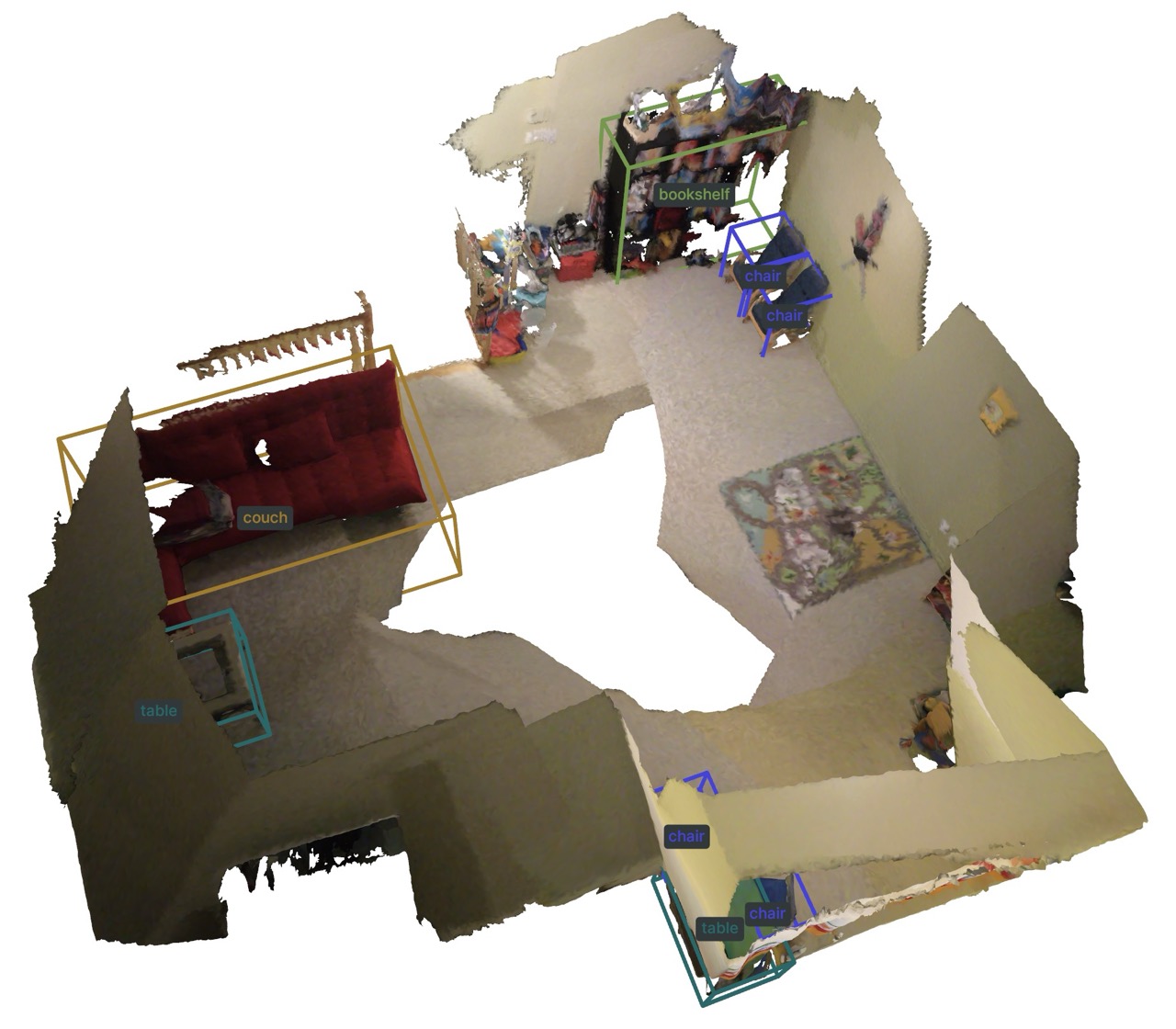}} &
   \raisebox{-0.5\height}{\includegraphics[width=0.25\textwidth]{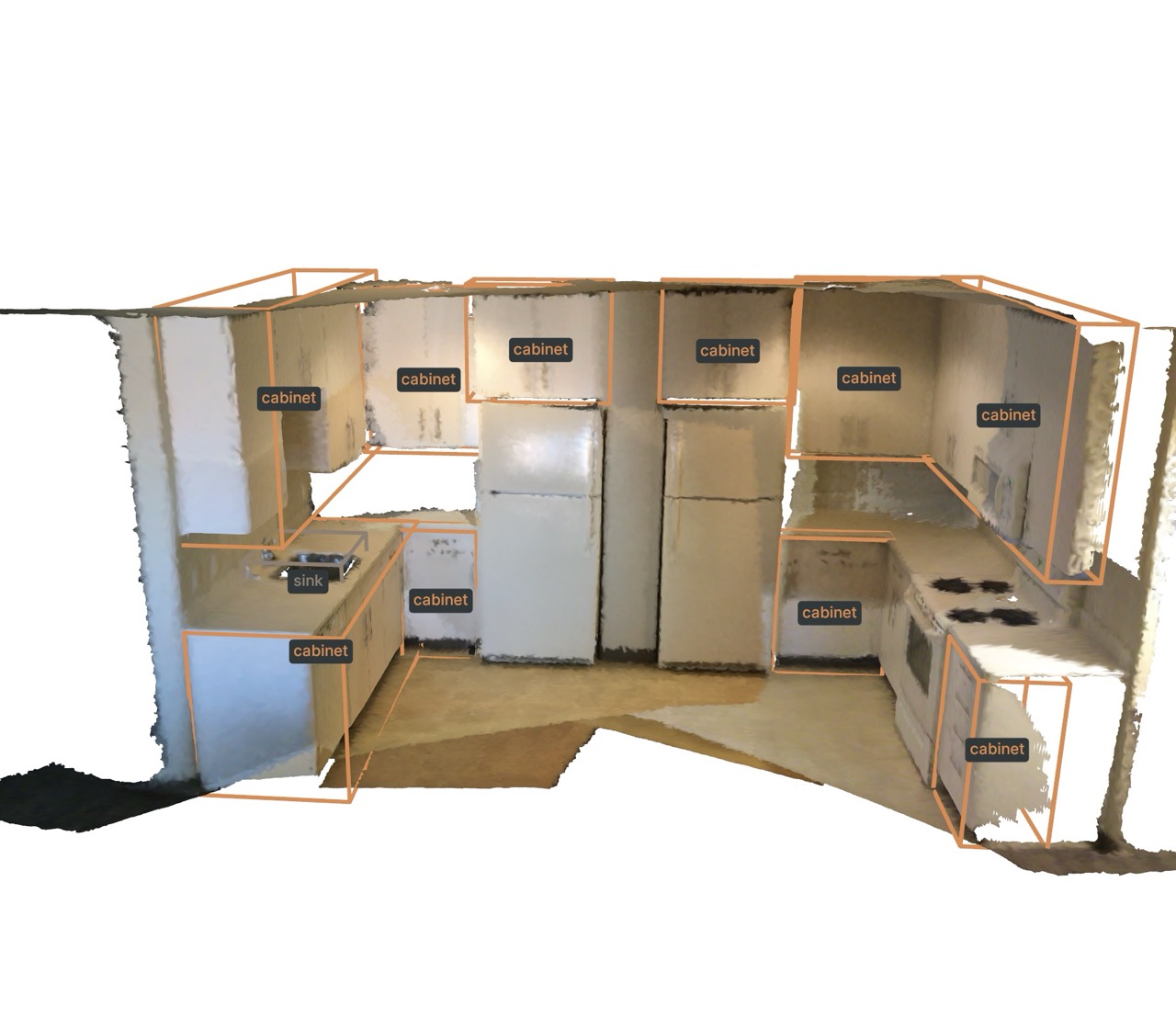}} &
   \raisebox{-0.5\height}{\includegraphics[width=0.25\textwidth]{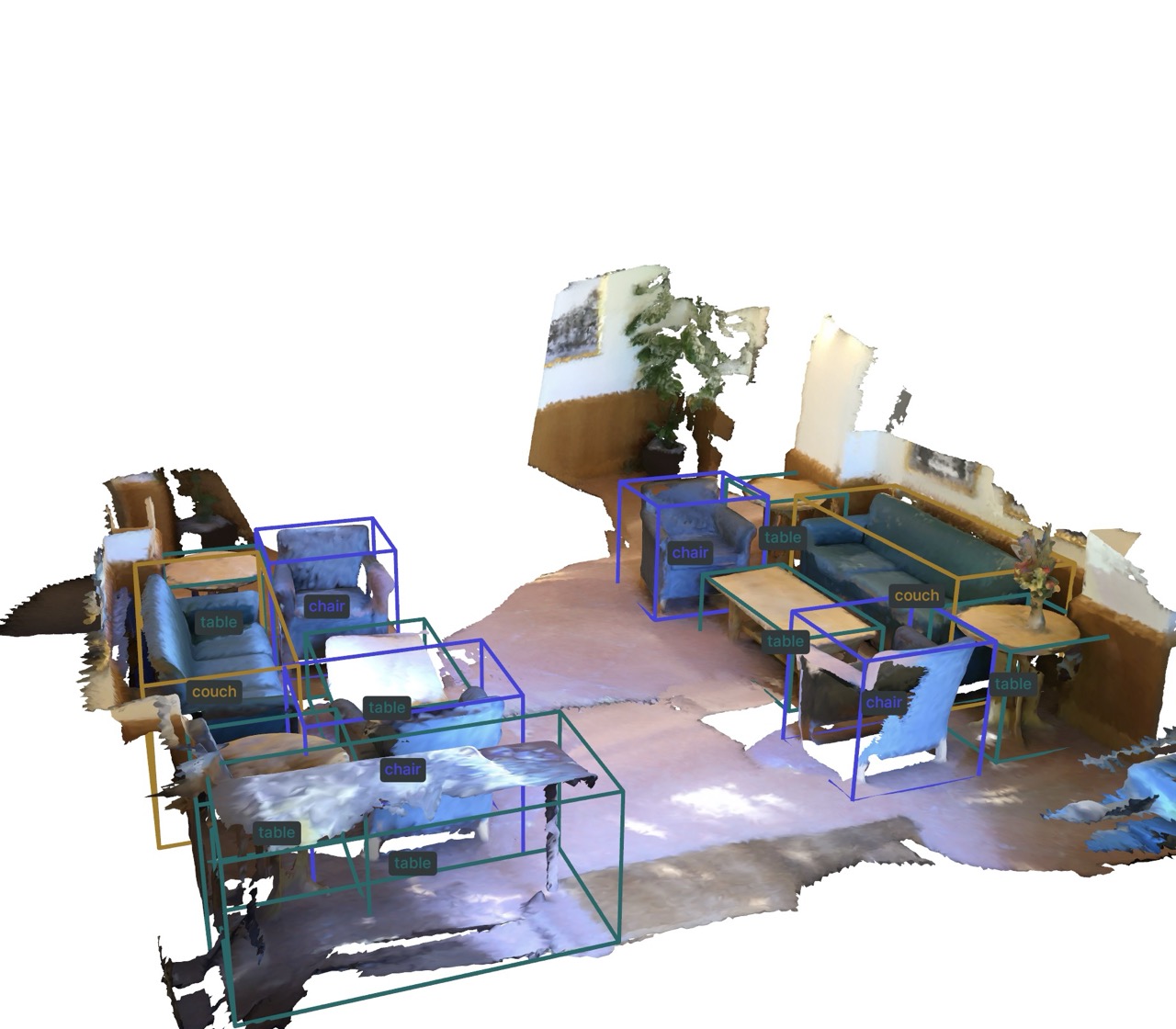}} &
   \raisebox{-0.5\height}{\includegraphics[width=0.25\textwidth]{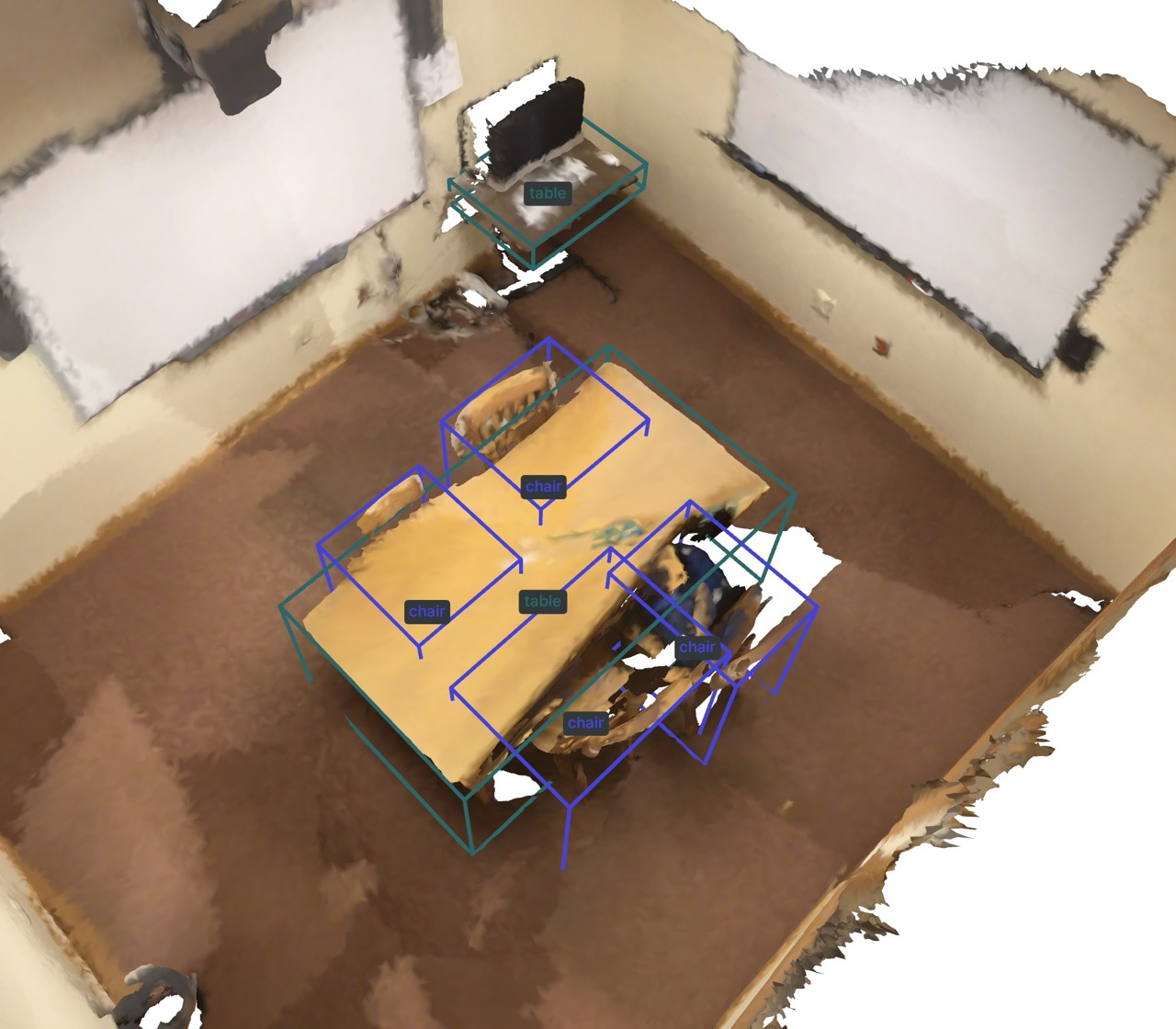}}
   \vspace{1mm}\\
   \rotatebox[origin=c]{90}{Prediction} &
   \raisebox{-0.5\height}{\includegraphics[width=0.25\textwidth]{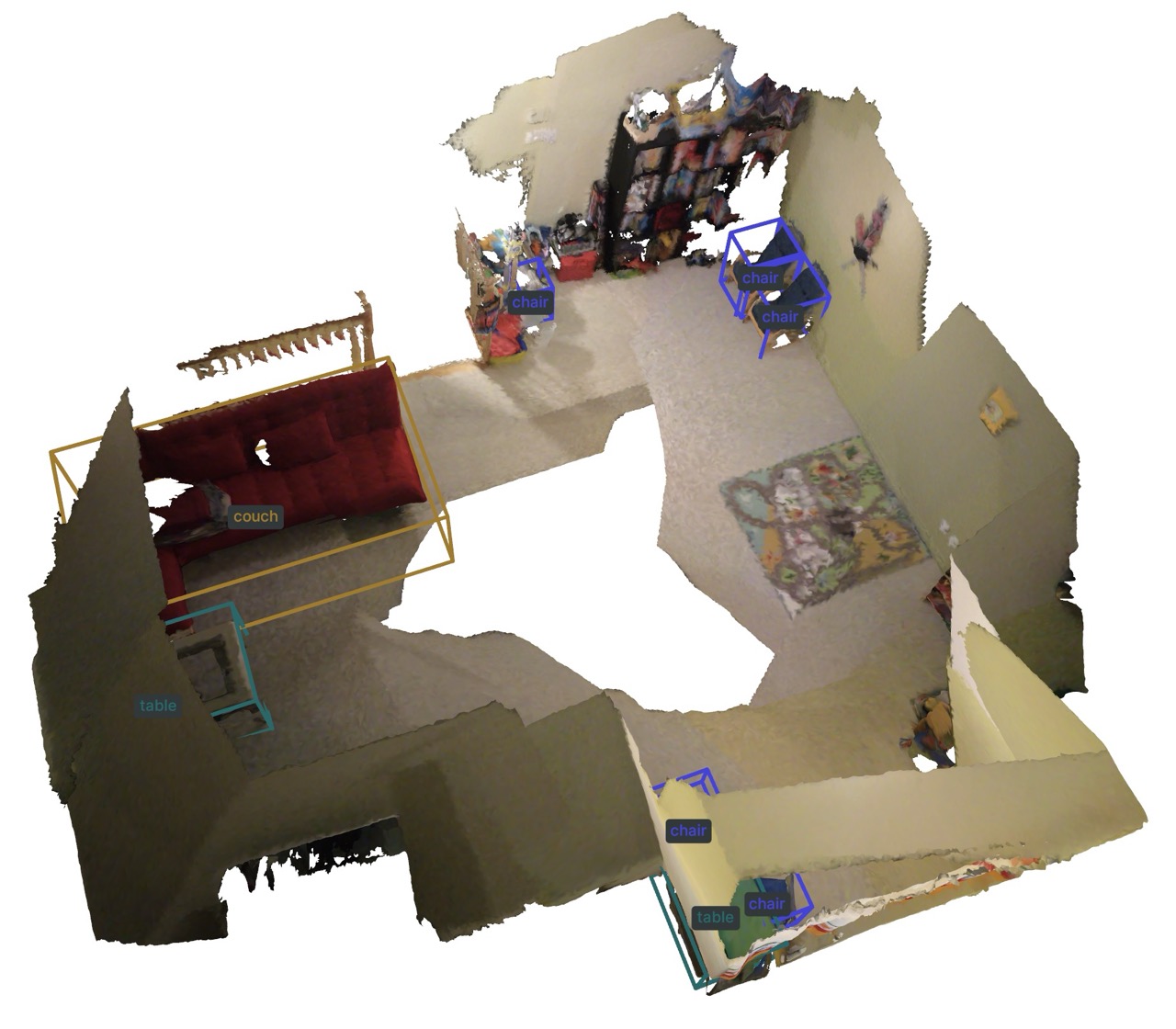}} &
   \raisebox{-0.5\height}{\includegraphics[width=0.25\textwidth]{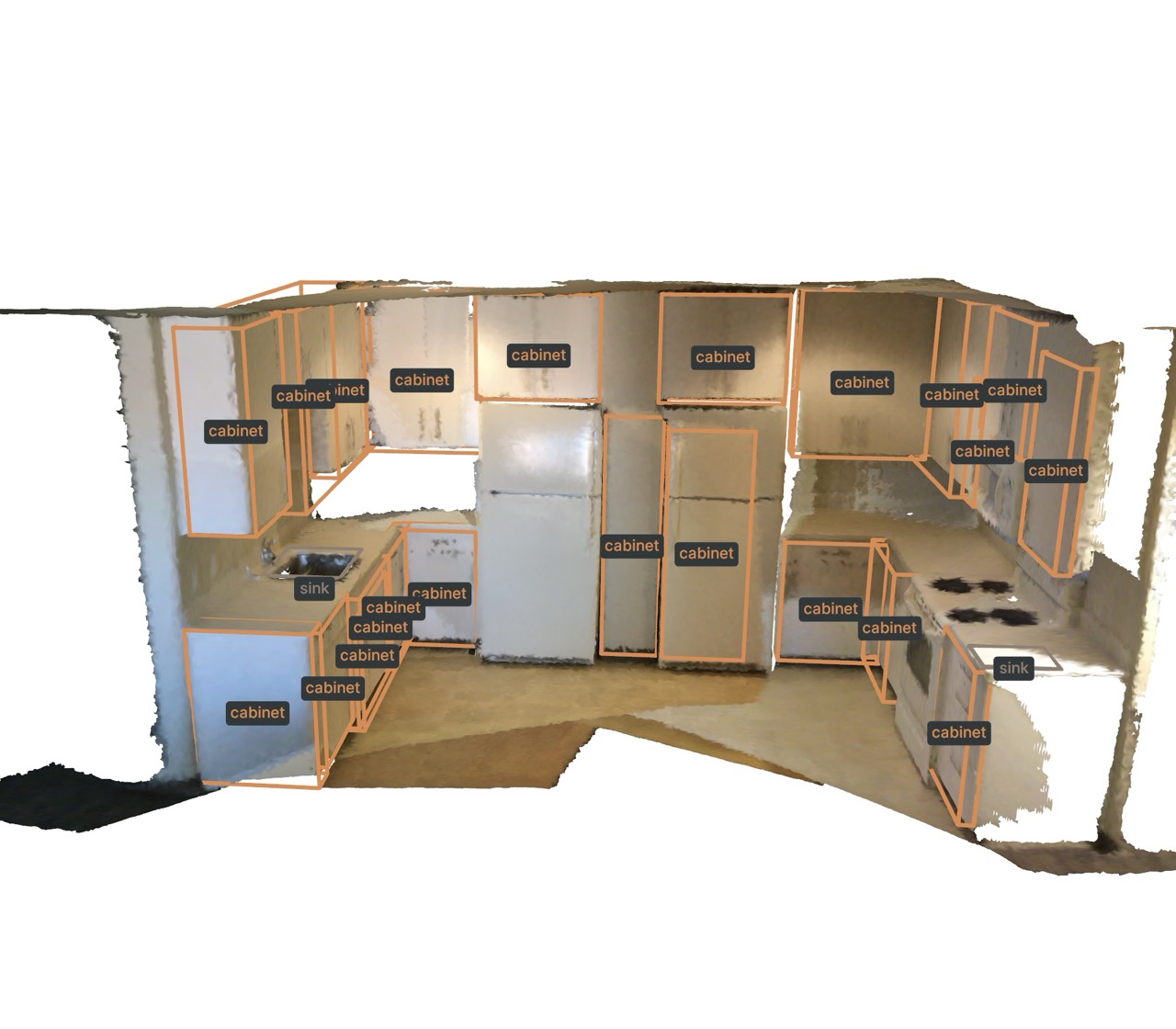}} &
   \raisebox{-0.5\height}{\includegraphics[width=0.25\textwidth]{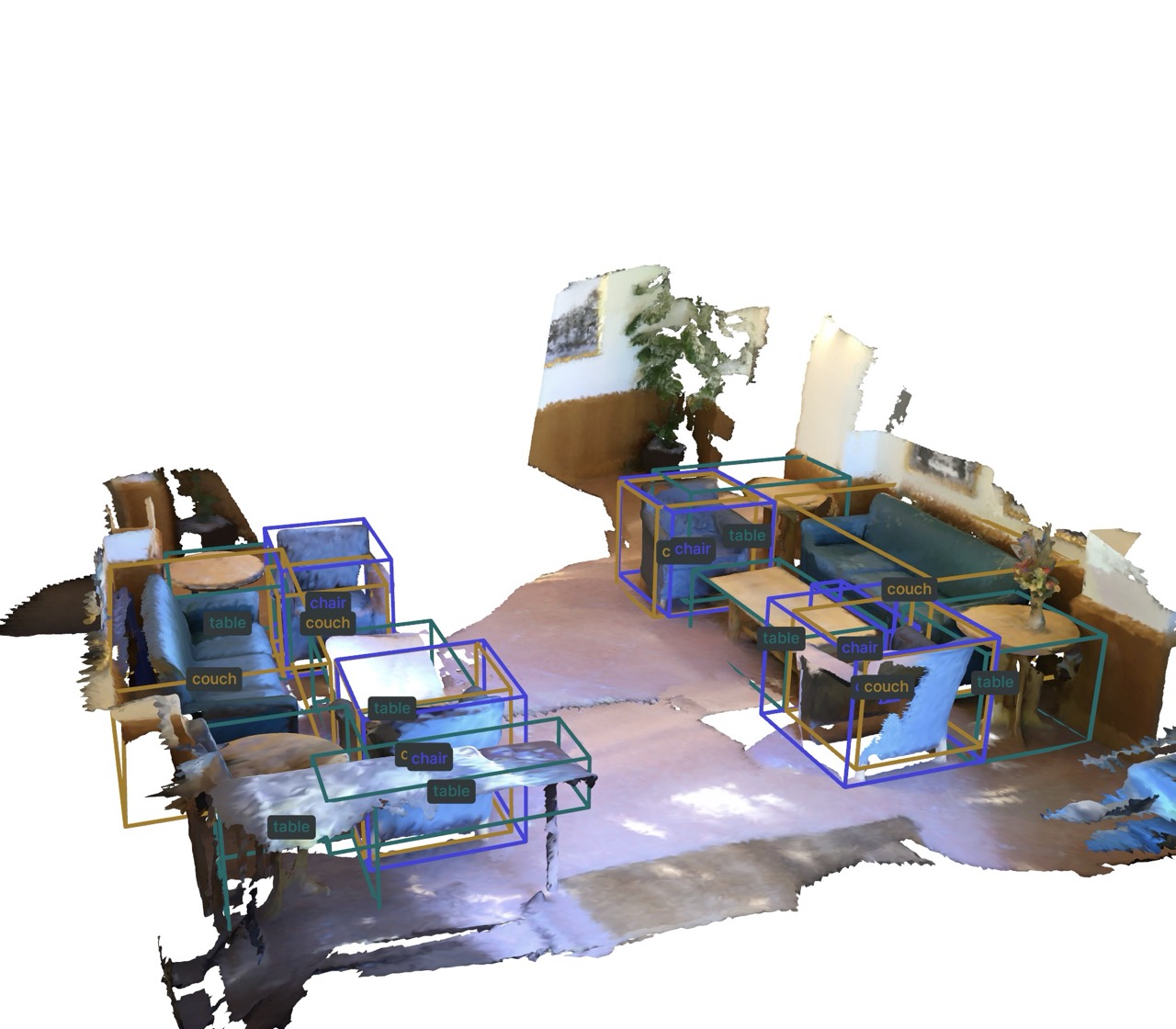}} &
   \raisebox{-0.5\height}{\includegraphics[width=0.25\textwidth]{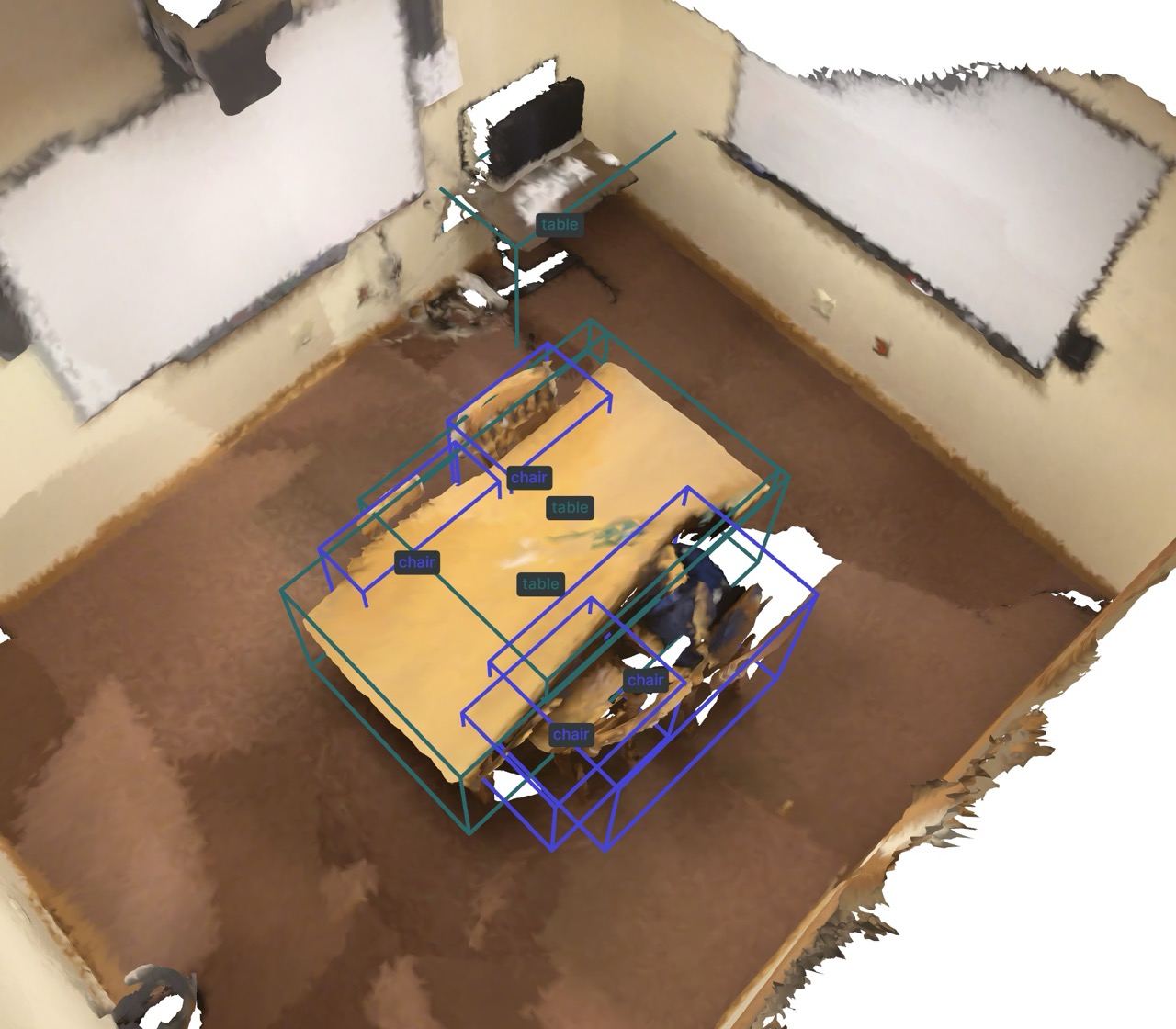}}
  \vspace{1mm}\\
   \rotatebox[origin=c]{90}{Ground truth} &
   \raisebox{-0.5\height}{\includegraphics[width=0.25\textwidth]{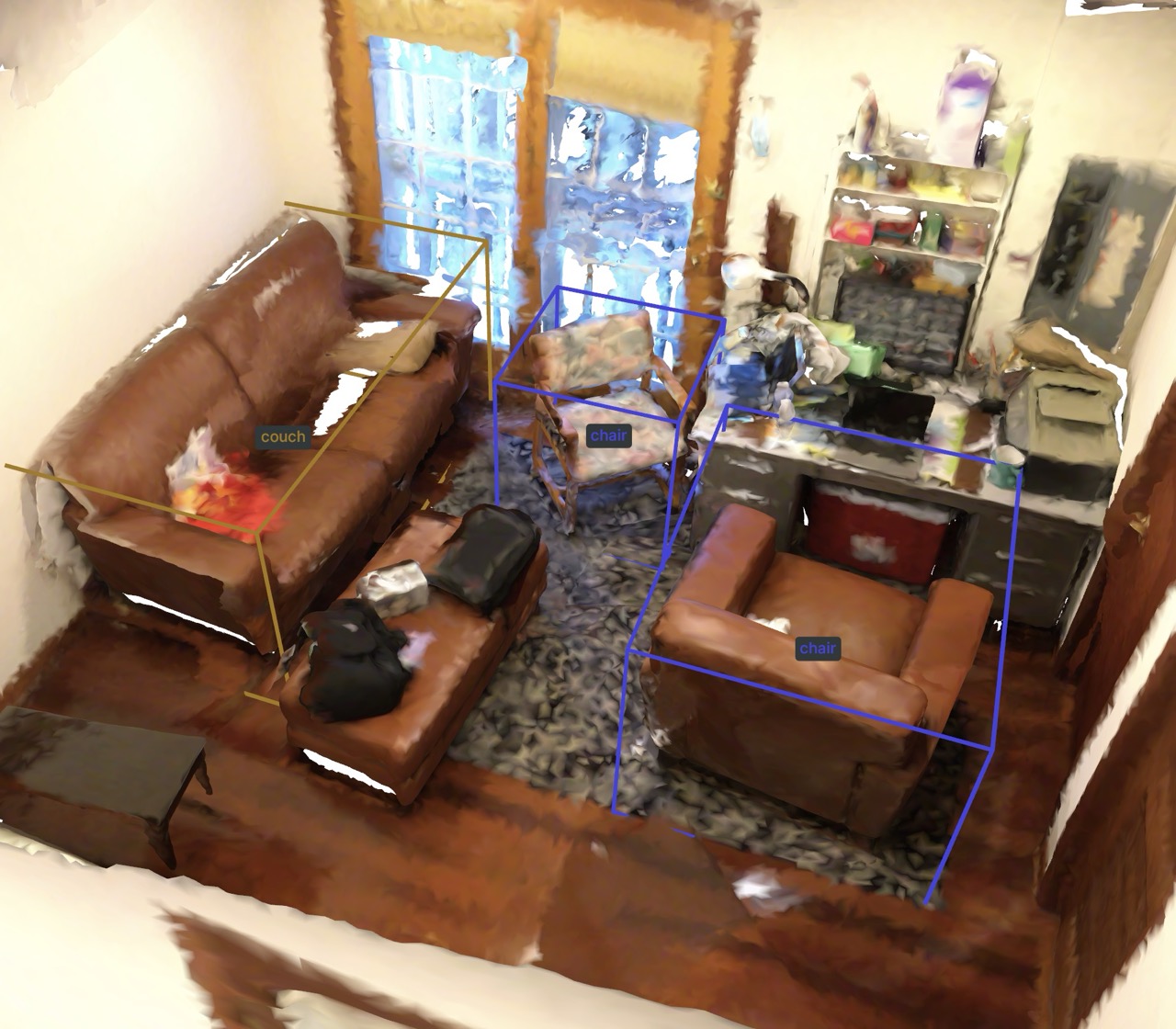}} &
   \raisebox{-0.5\height}{\includegraphics[width=0.25\textwidth]{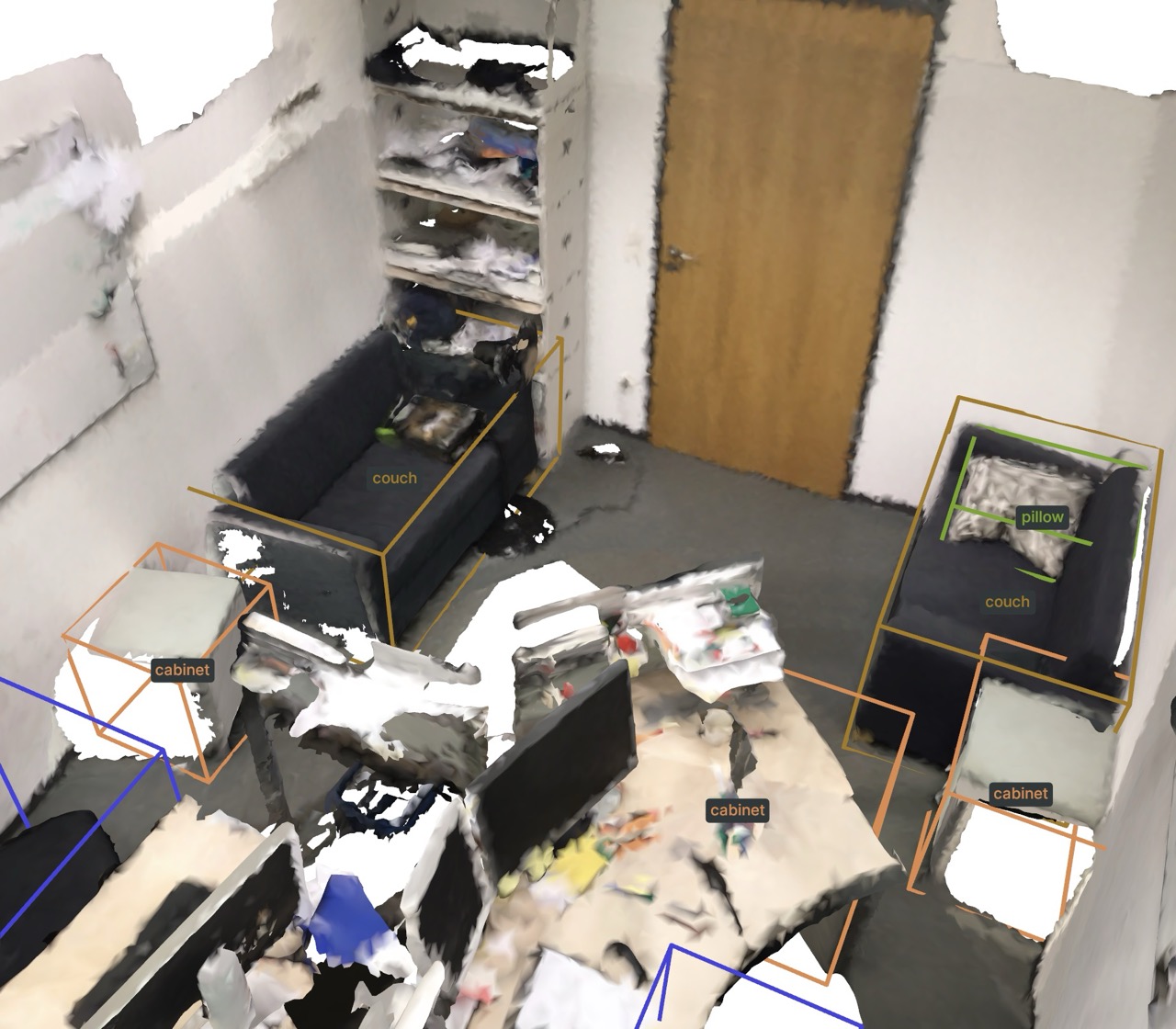}} &
   \raisebox{-0.5\height}{\includegraphics[width=0.25\textwidth]{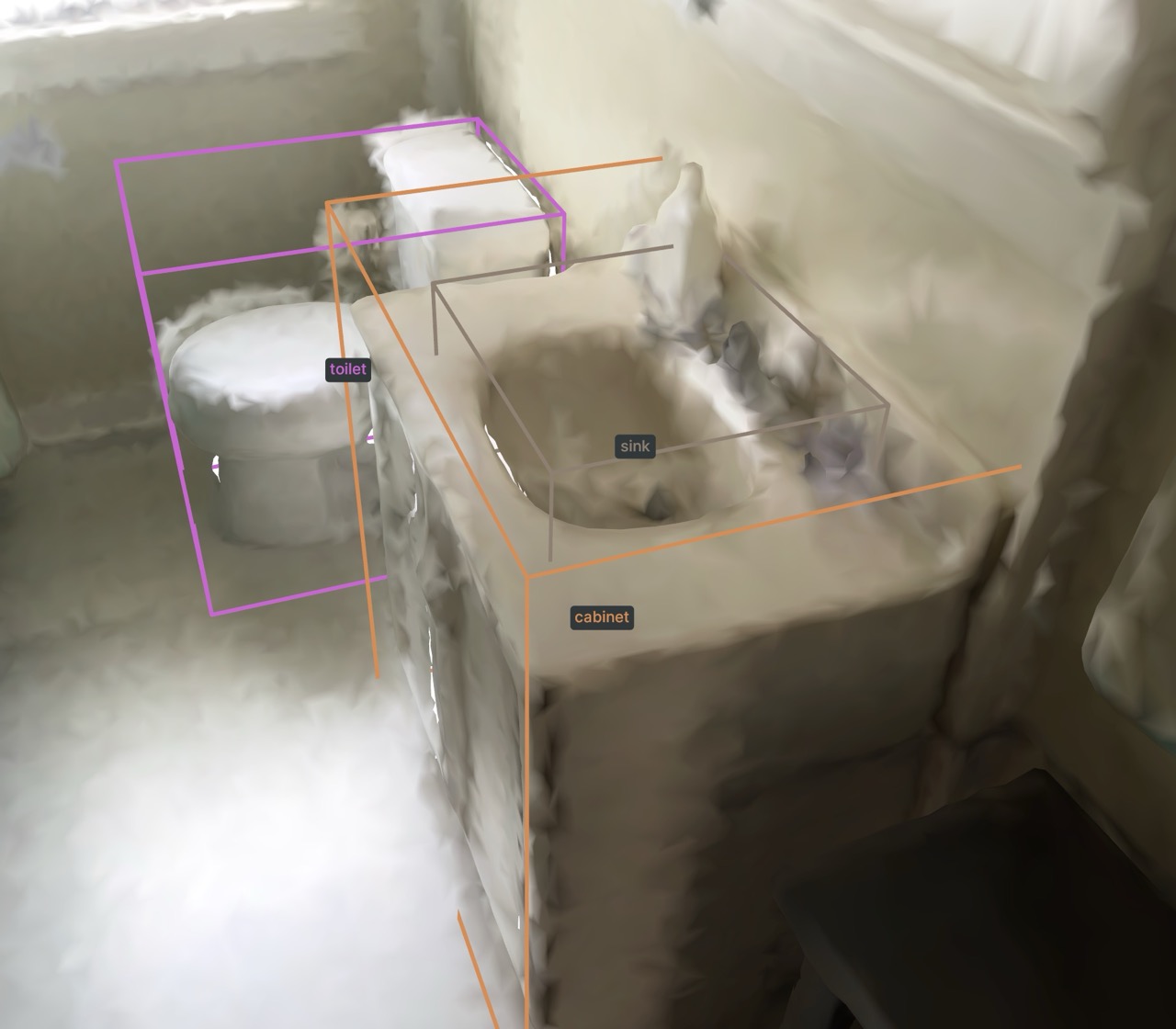}} &
   \raisebox{-0.5\height}{\includegraphics[width=0.25\textwidth]{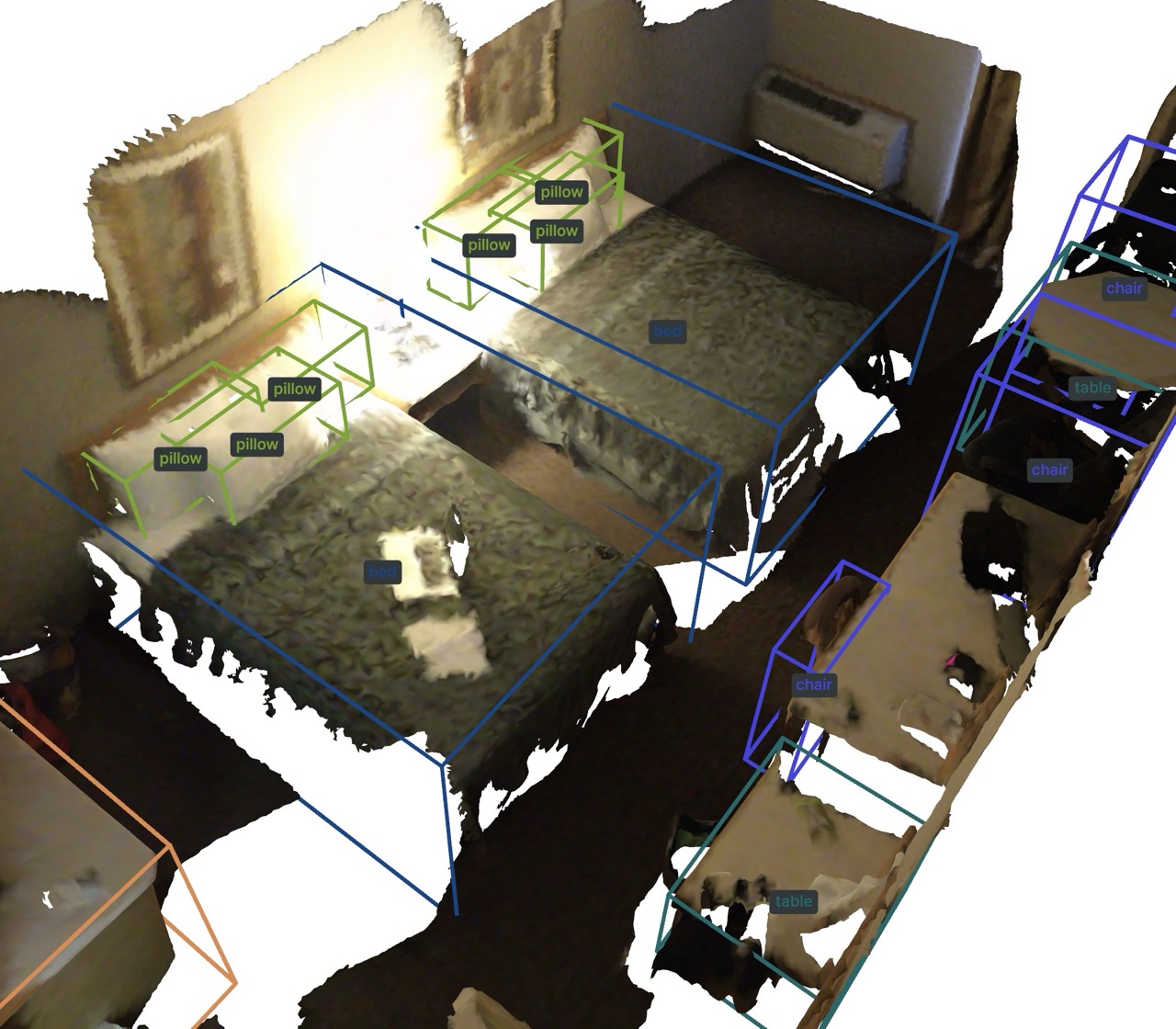}}
  \vspace{1mm}\\
   \rotatebox[origin=c]{90}{Prediction} &
   \raisebox{-0.5\height}{\includegraphics[width=0.25\textwidth]{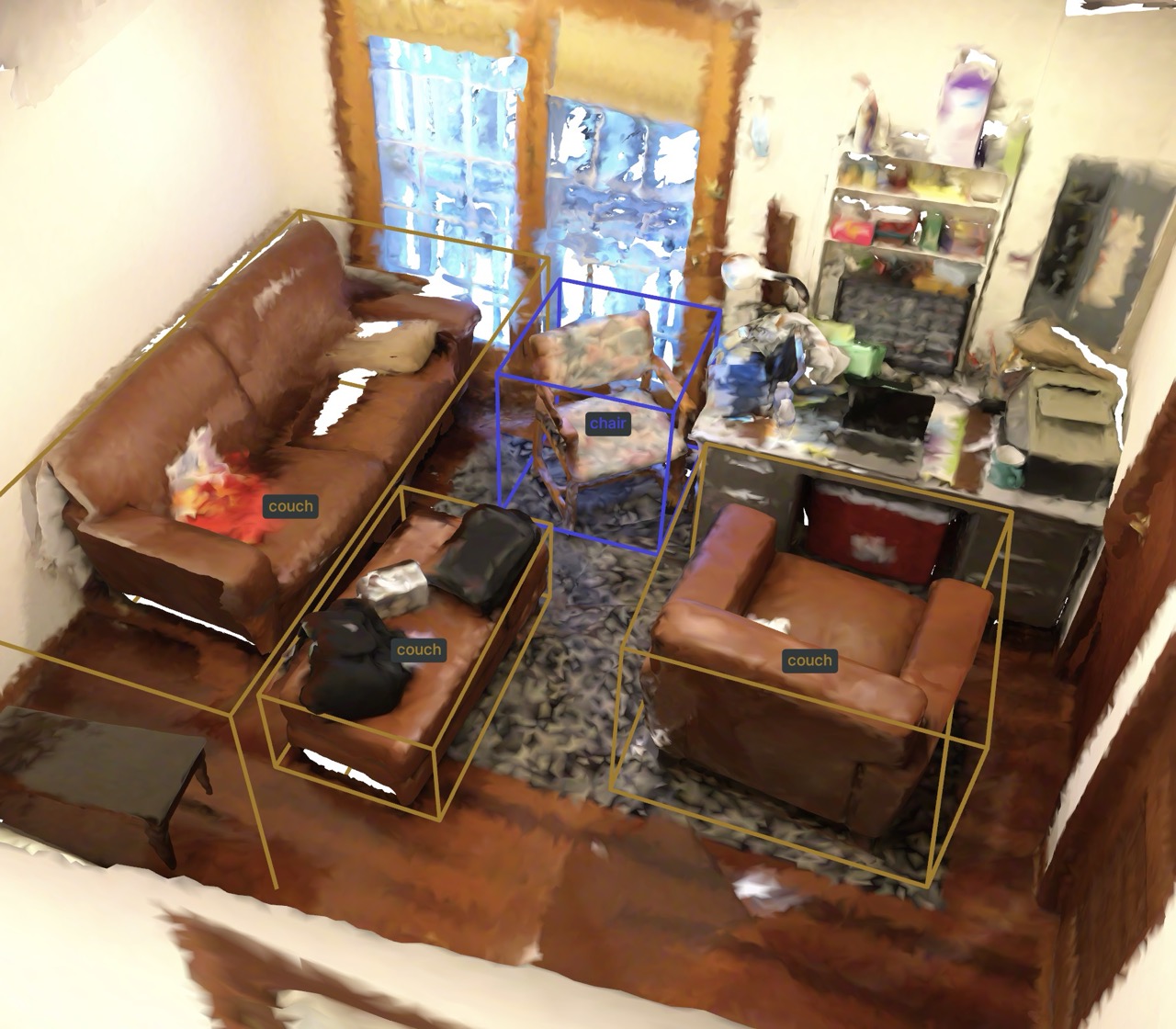}} &
   \raisebox{-0.5\height}{\includegraphics[width=0.25\textwidth]{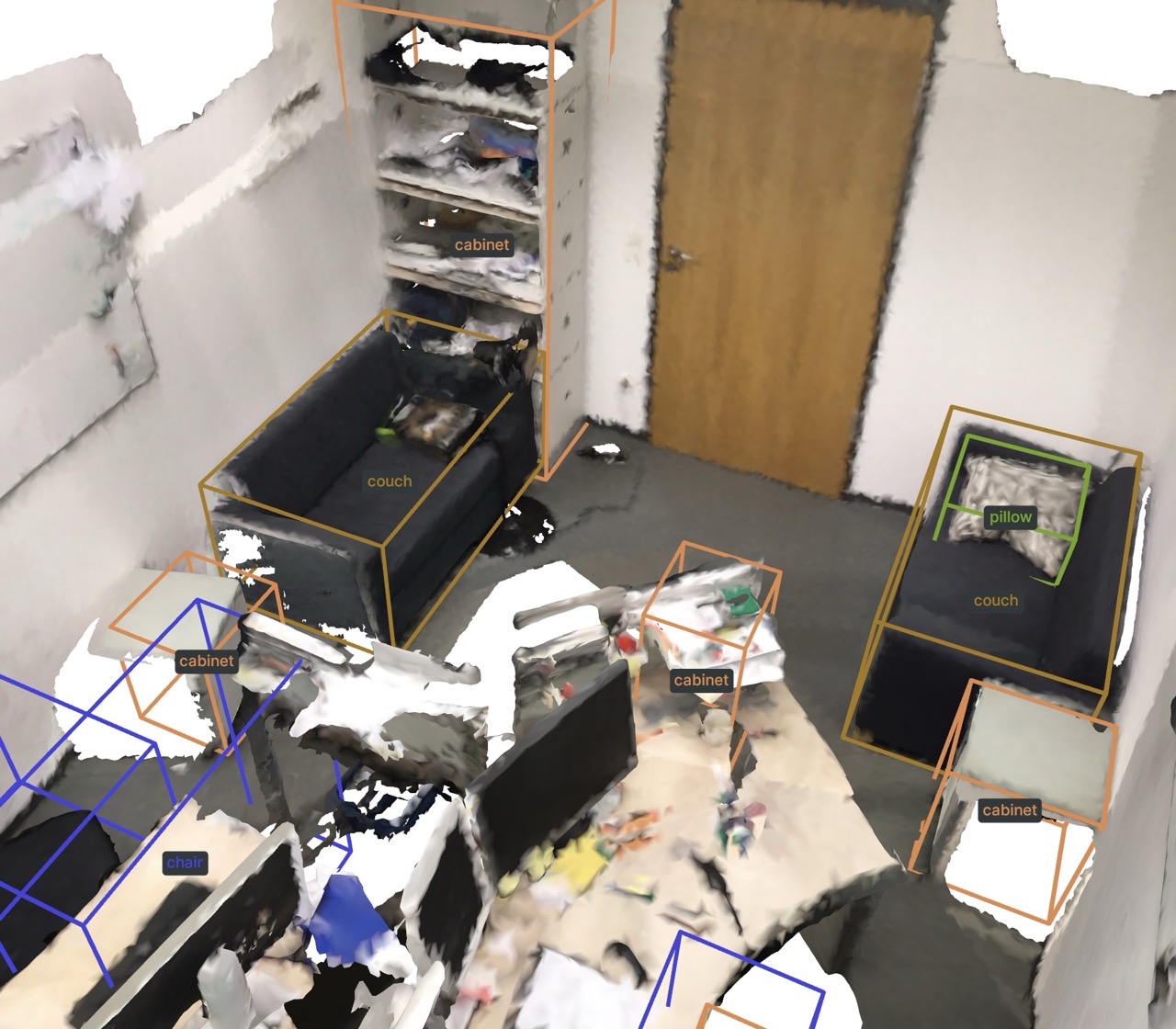}} &
   \raisebox{-0.5\height}{\includegraphics[width=0.25\textwidth]{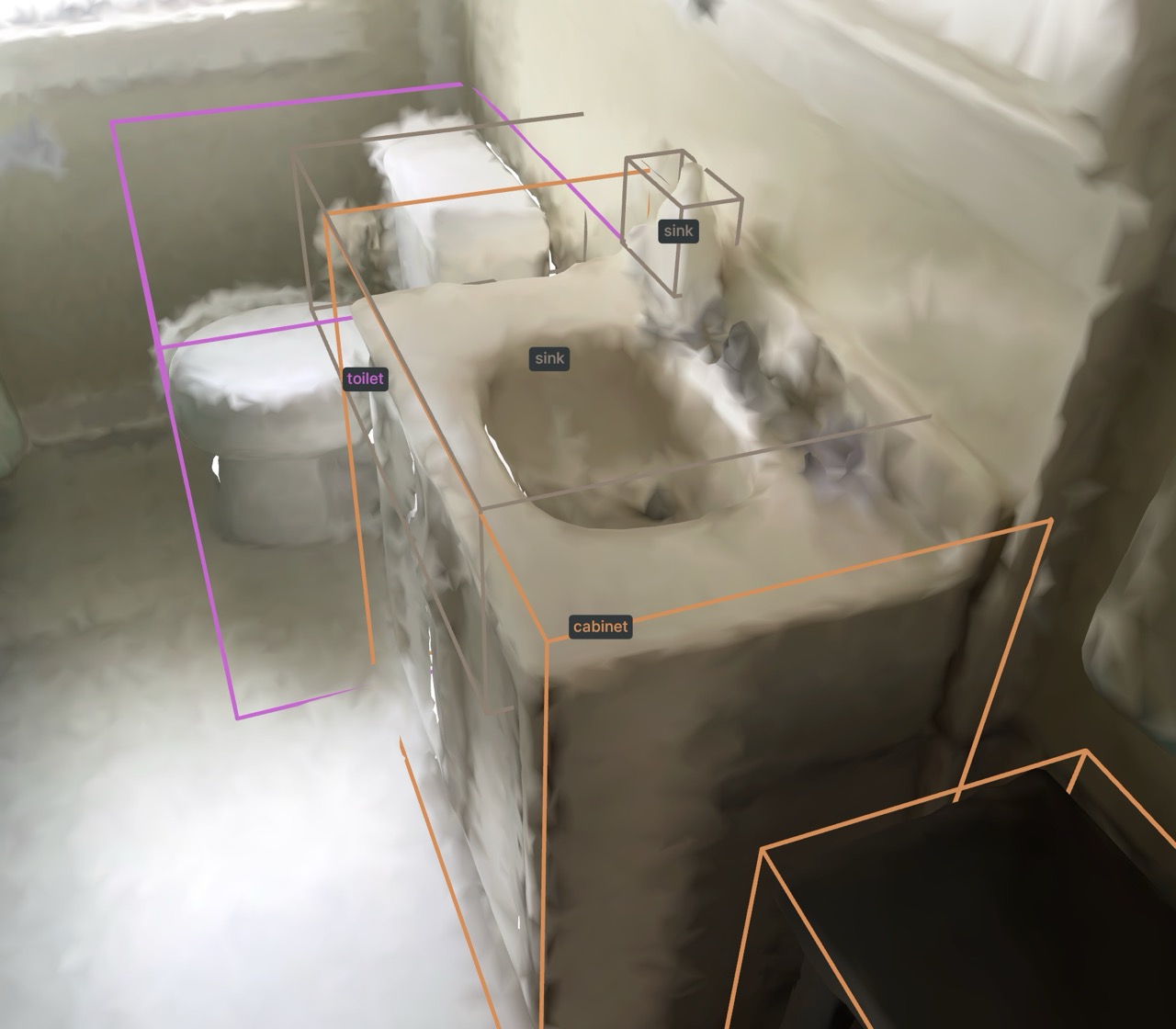}} &
   \raisebox{-0.5\height}{\includegraphics[width=0.25\textwidth]{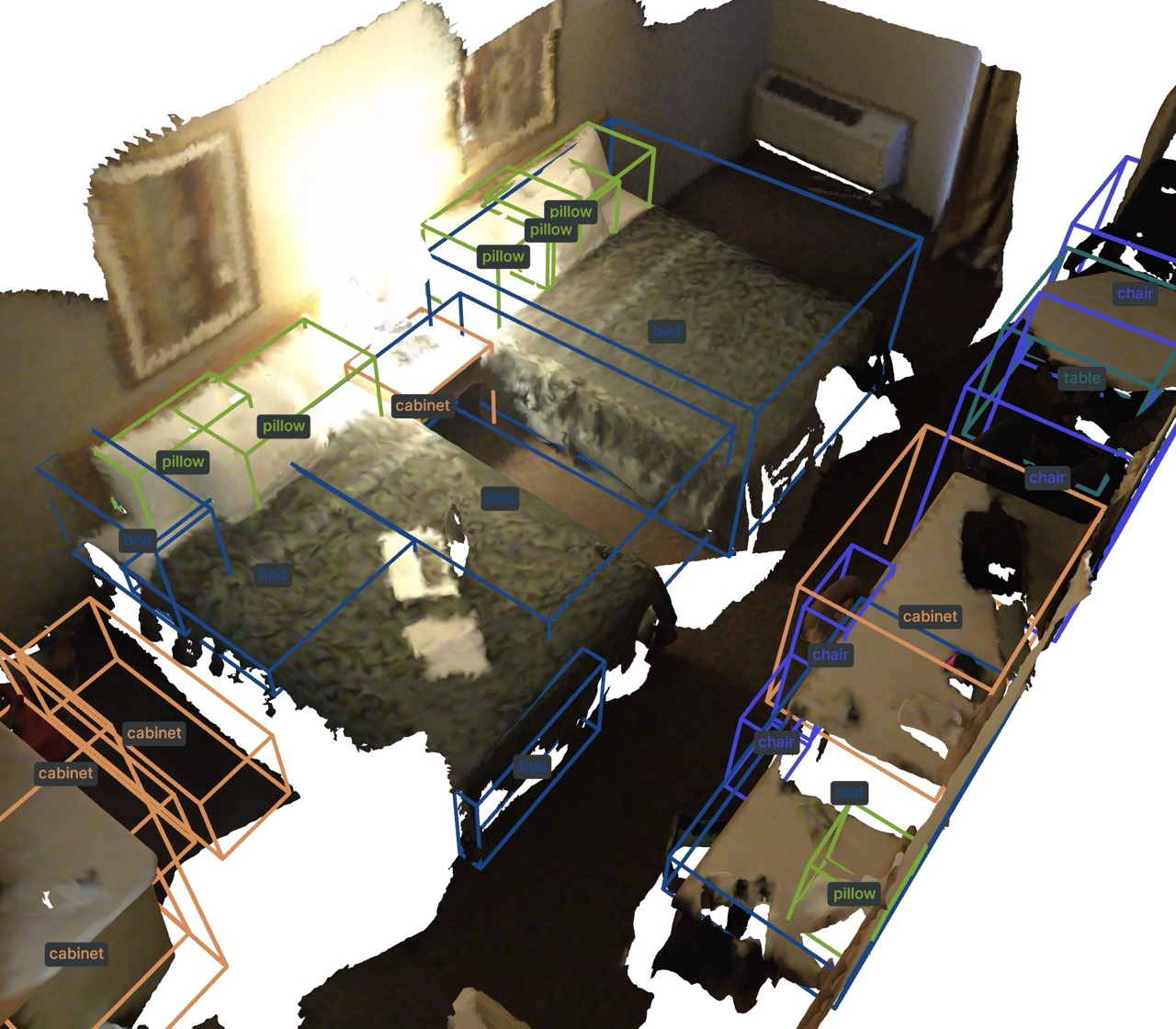}}
 \end{tabular}
 }
\caption{\textbf{Qualitative results on ScanNet.} Zero-shot 3D object detection, using the ScanNet10 categories as prompts. Best seen on screen.}
\label{fig:qualitative}
\vspace{-6pt}
\end{figure*}

%% file: tables/tab_scannet_200.tex
\begin{table}[t]
\small
\centering
\begin{tabular}[t]{l ccc}
\toprule[1pt]
Method & Head & Common & Tail\T\B\\
\midrule[.5pt]
Object2Scene~\cite{zhu2023object2scene} & - & 10.1 &  3.4\T\\
OpenIns3D~\cite{huang2024openins3d} \tiny{with RGB-D} & \textbf{25.6} & 20.4 & 16.5\\
\OURS (Ours)  \tiny{RGB-D}& 23.2 &  \textbf{25.9} & \textbf{33.2}\B\\ 
\midrule[.2pt]
\OURS (Ours) & 13.3 &  18.2 & 16.7\B\T\\ 
\bottomrule[1pt]
\end{tabular}
\caption{\label{table:scannet_200}\textbf{Open-vocabulary Object Detection on ScanNet 200.} We use the whole ScanNet200 vocabulary as prompt (except \textit{'wall'} and \textit{'floor'}), and present results (mAP$_{25}$) for the \textit{Head}, \textit{Common} and \textit{Tail} category splits.
Our method performs just as well on long-tail classes (Tail) as on the most frequent ones (Head).
}
\end{table}

%% file: figures/fig_ablation_nviews.tex
\begin{figure}[t]
\centering
\includegraphics[width=0.9\linewidth]{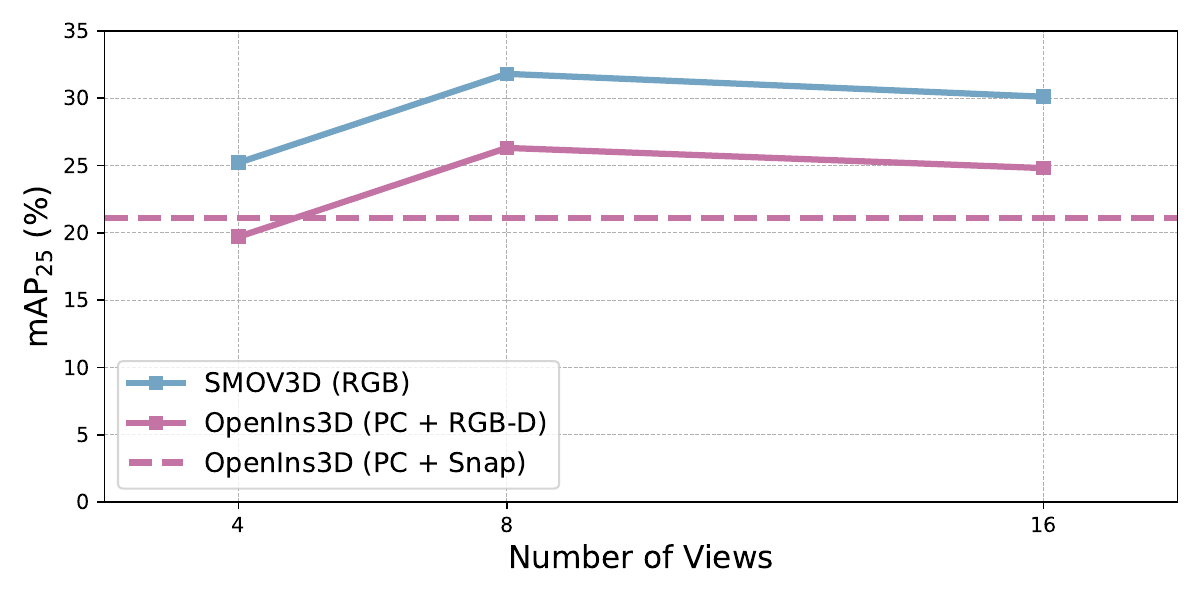}
\caption{\label{tab:ablation_nviews}\textbf{Few views.} We report mAP$_{25}$ scores (\%) on the Replica dataset with few views chosen so that the entire scene is visible, and compare to OpenIns3D.}
\end{figure}

%% file: figures/arkitscenes.tex
\begin{figure*}[t!]
\scriptsize
\centering
 \begin{tabular}{@{}c@{\hspace{1mm}}c@{\hspace{1mm}}c@{\hspace{1mm}}c@{\hspace{1mm}}c@{}}
   \raisebox{-0.5\height}{\includegraphics[width=0.25\textwidth]{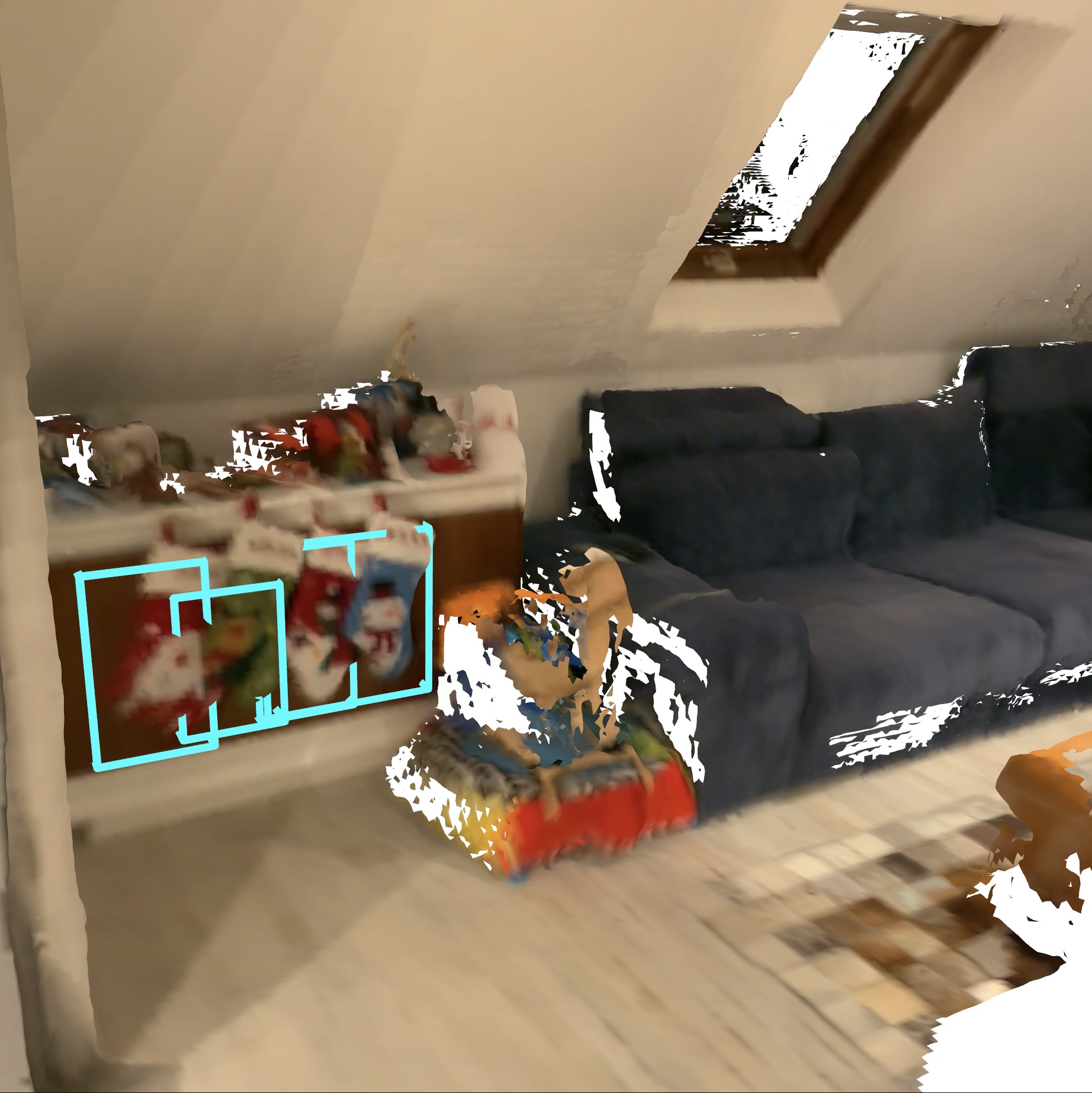}} &
   \raisebox{-0.5\height}{\includegraphics[width=0.25\textwidth]{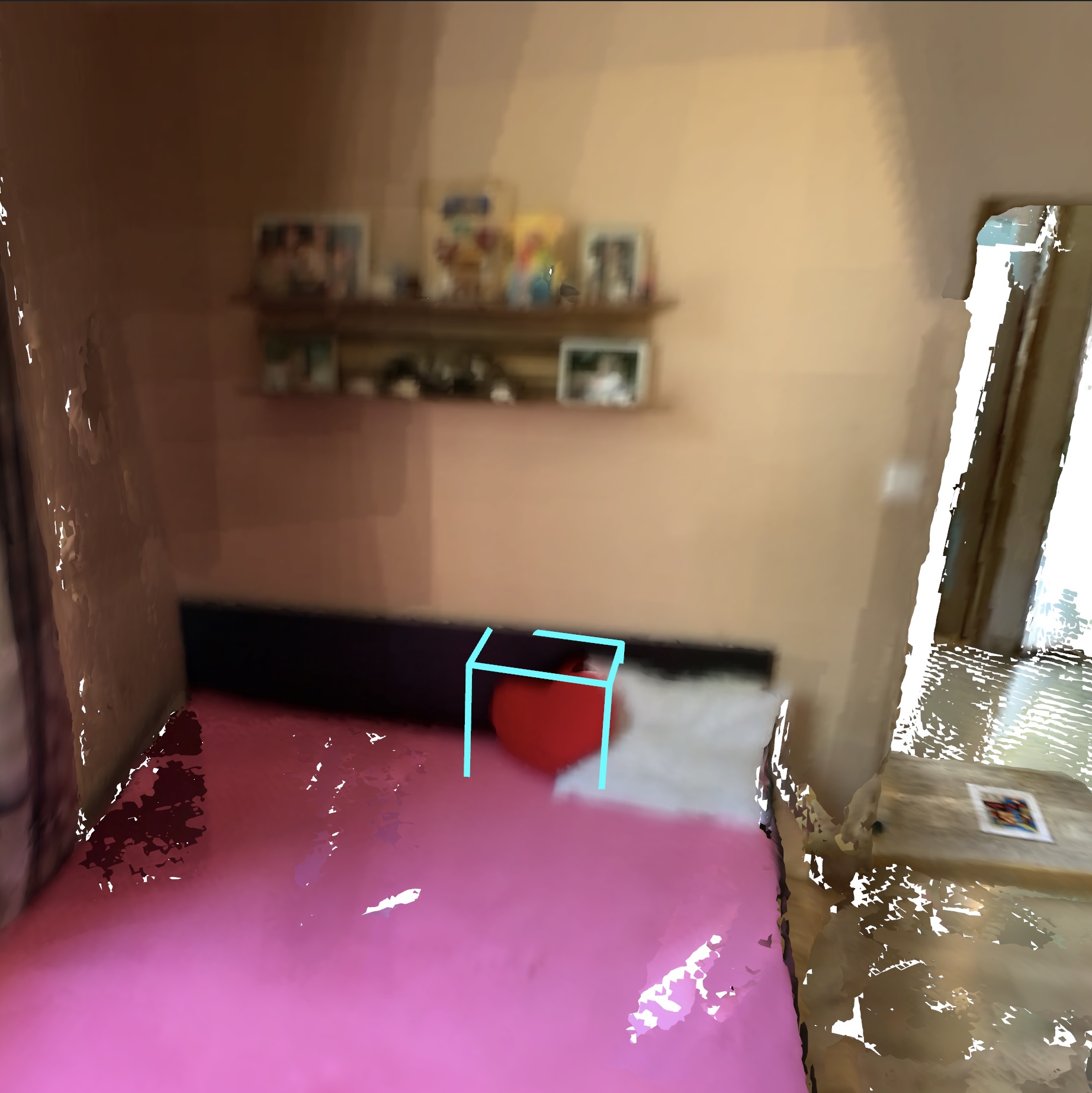}} &
   \raisebox{-0.5\height}{\includegraphics[width=0.25\textwidth]{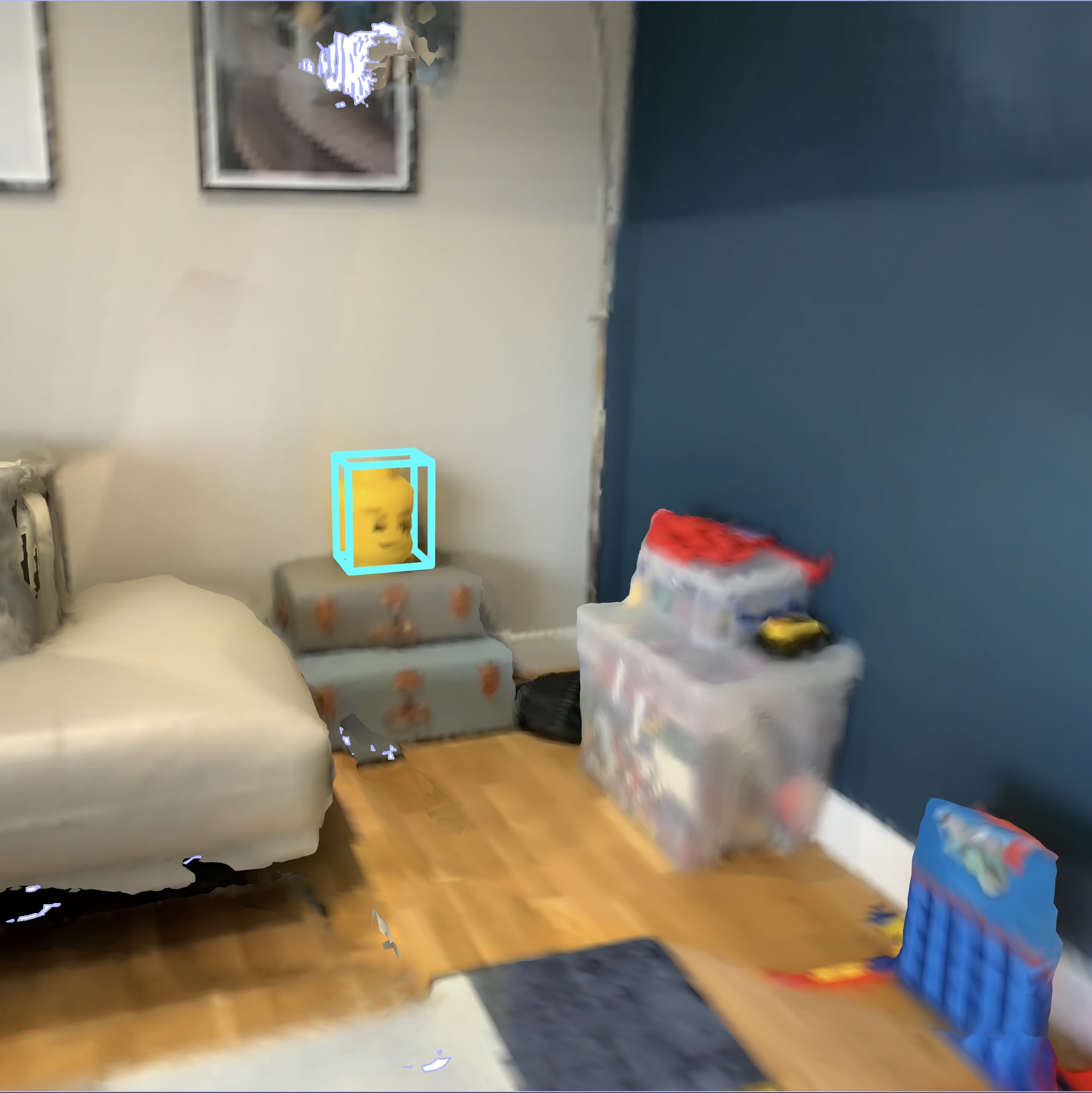}} &
   \raisebox{-0.5\height}{\includegraphics[width=0.25\textwidth]{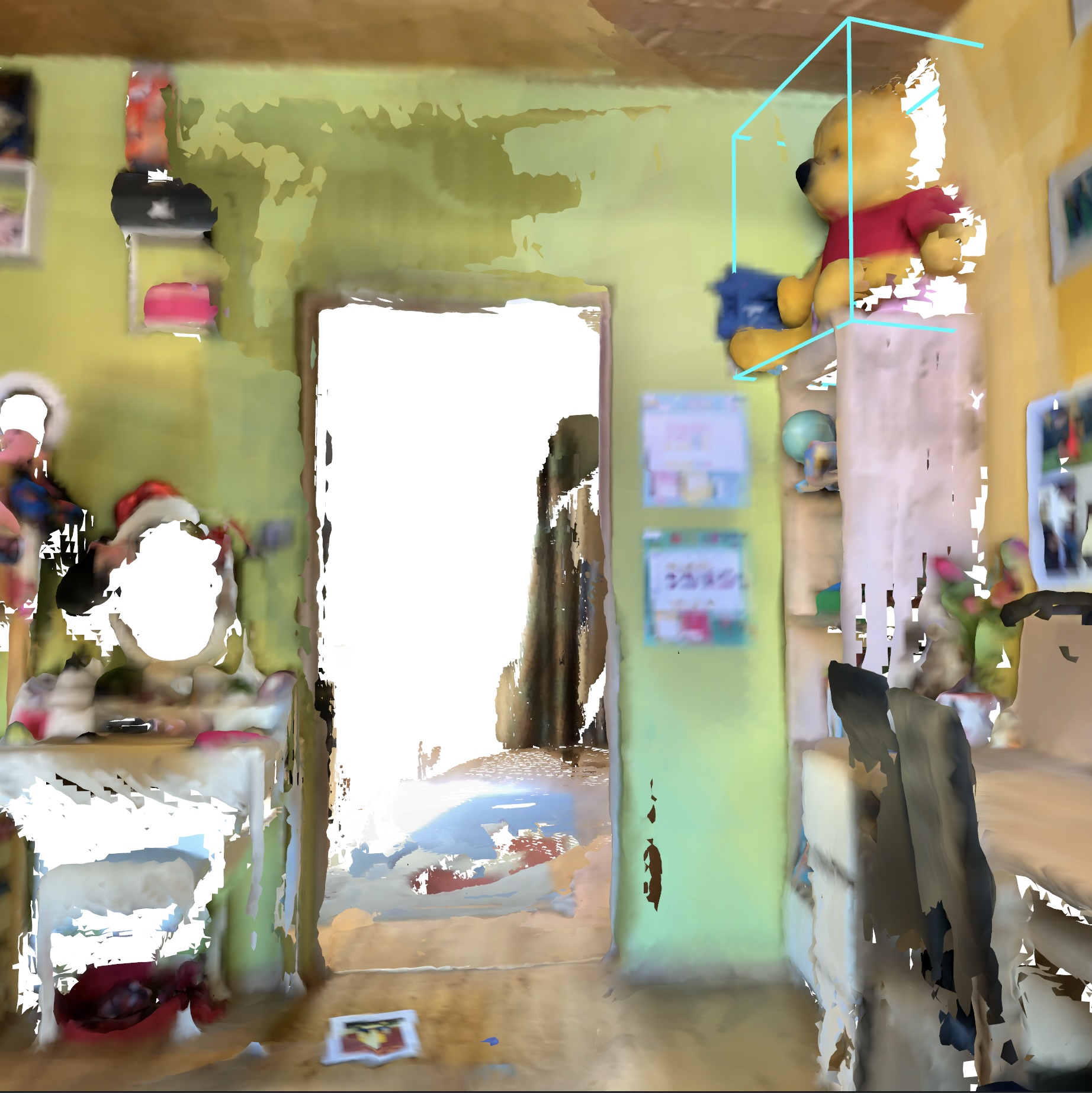}}
   \vspace{1mm}\\
   "Christmas stockings" &
   "Heart-shaped pillow" &
   "Lego head" &
   "Winnie-the-Pooh"
 \end{tabular}
\caption{\textbf{Qualitative results on ARKitScenes.} Zero-shot 3D object detection, using the prompts proposed in the OpenSUN3D challenge.}
\label{fig:arkitscenes}

%% file: figures/replica.tex
\scriptsize
\centering
 \begin{tabular}{@{}c@{\hspace{1mm}}c@{\hspace{1mm}}c@{\hspace{1mm}}c@{\hspace{1mm}}c@{}}
   \rotatebox[origin=c]{90}{room0, 4 views} &
   \raisebox{-0.5\height}{\includegraphics[width=0.33\textwidth]{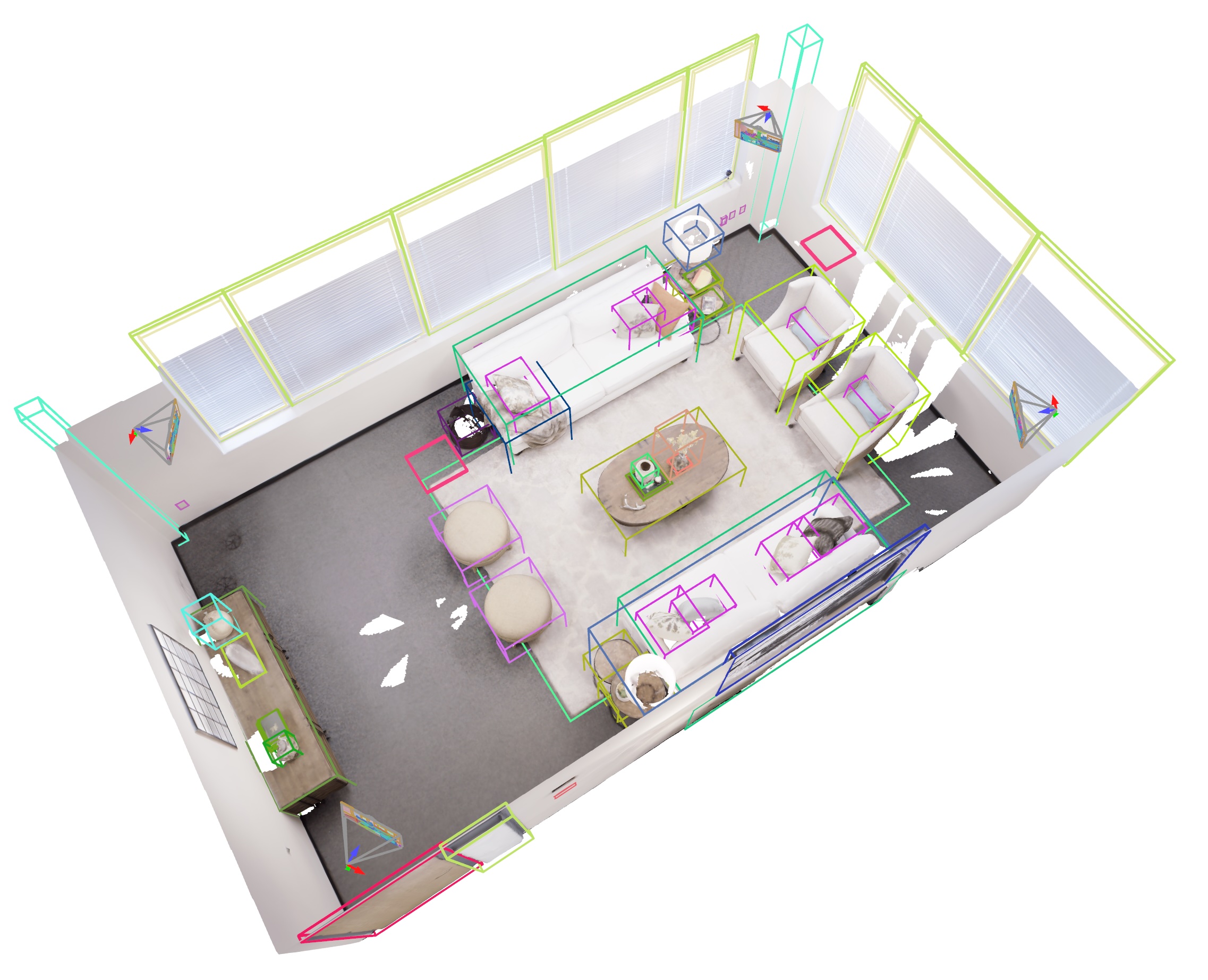}} &
   \raisebox{-0.5\height}{\includegraphics[width=0.33\textwidth]{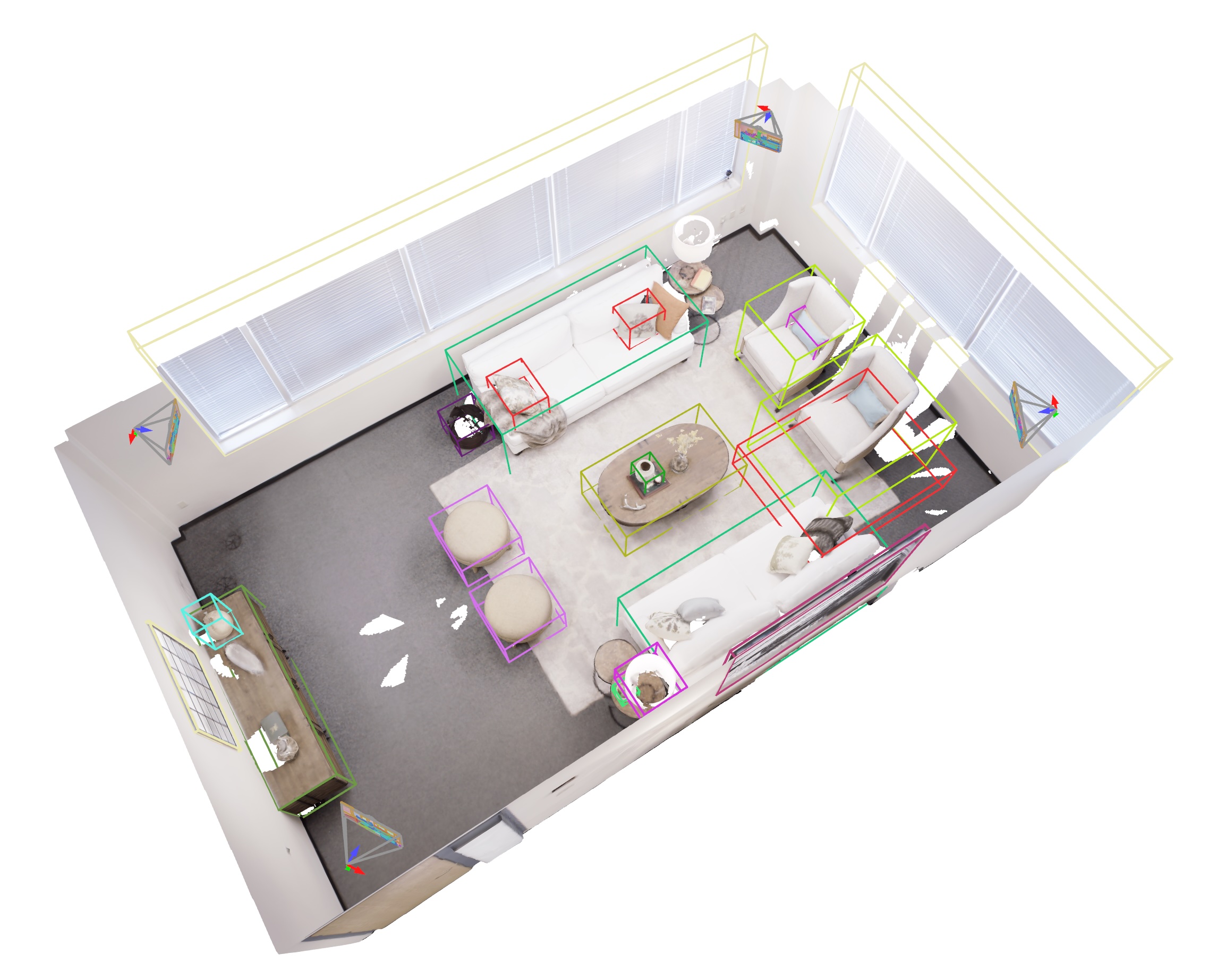}} &
   \raisebox{-0.5\height}{\includegraphics[width=0.33\textwidth]{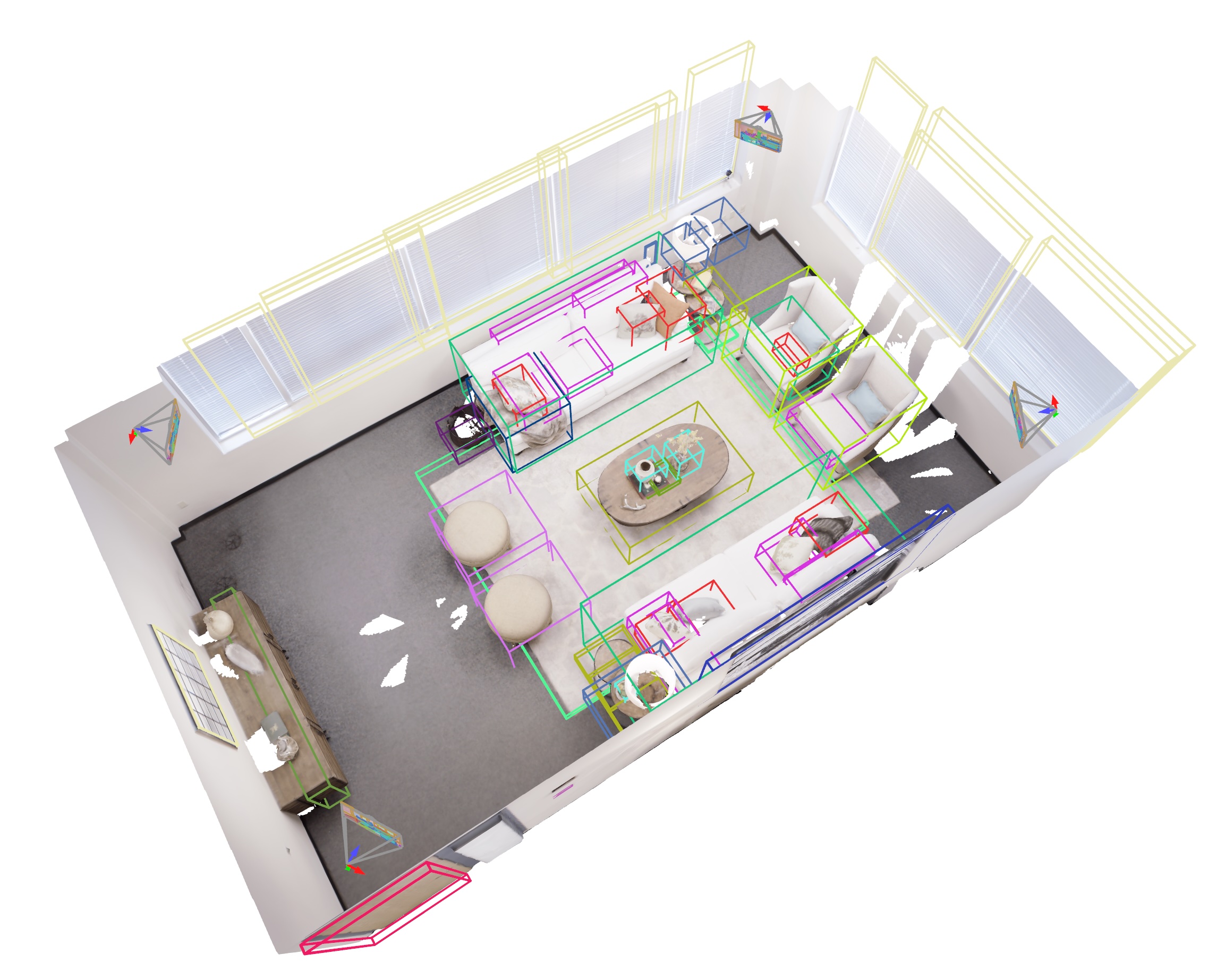}}
   \vspace{1mm}\\
   \rotatebox[origin=c]{90}{office2, 8 views} &
   \raisebox{-0.5\height}{\includegraphics[width=0.33\textwidth]{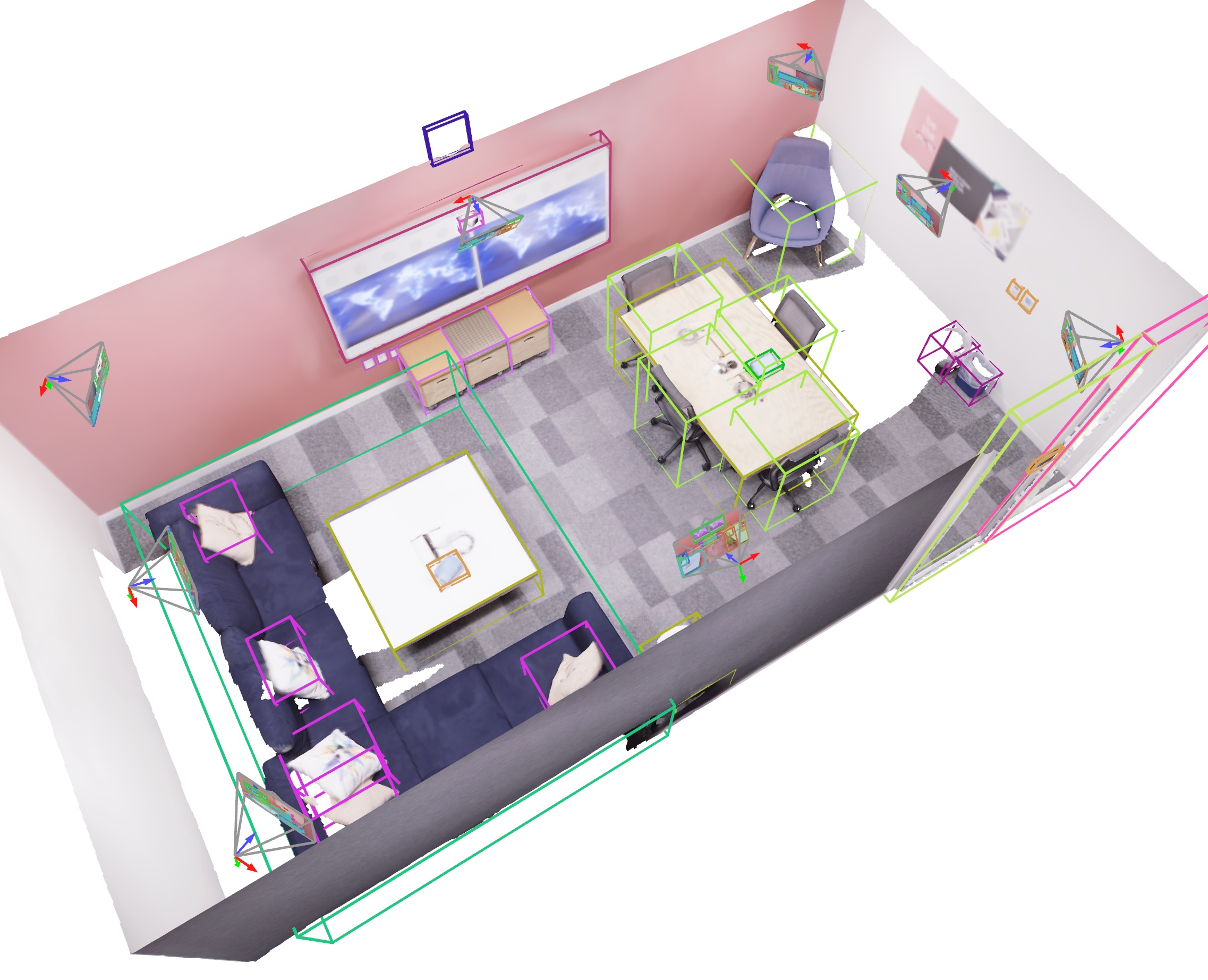}} &
   \raisebox{-0.5\height}{\includegraphics[width=0.33\textwidth]{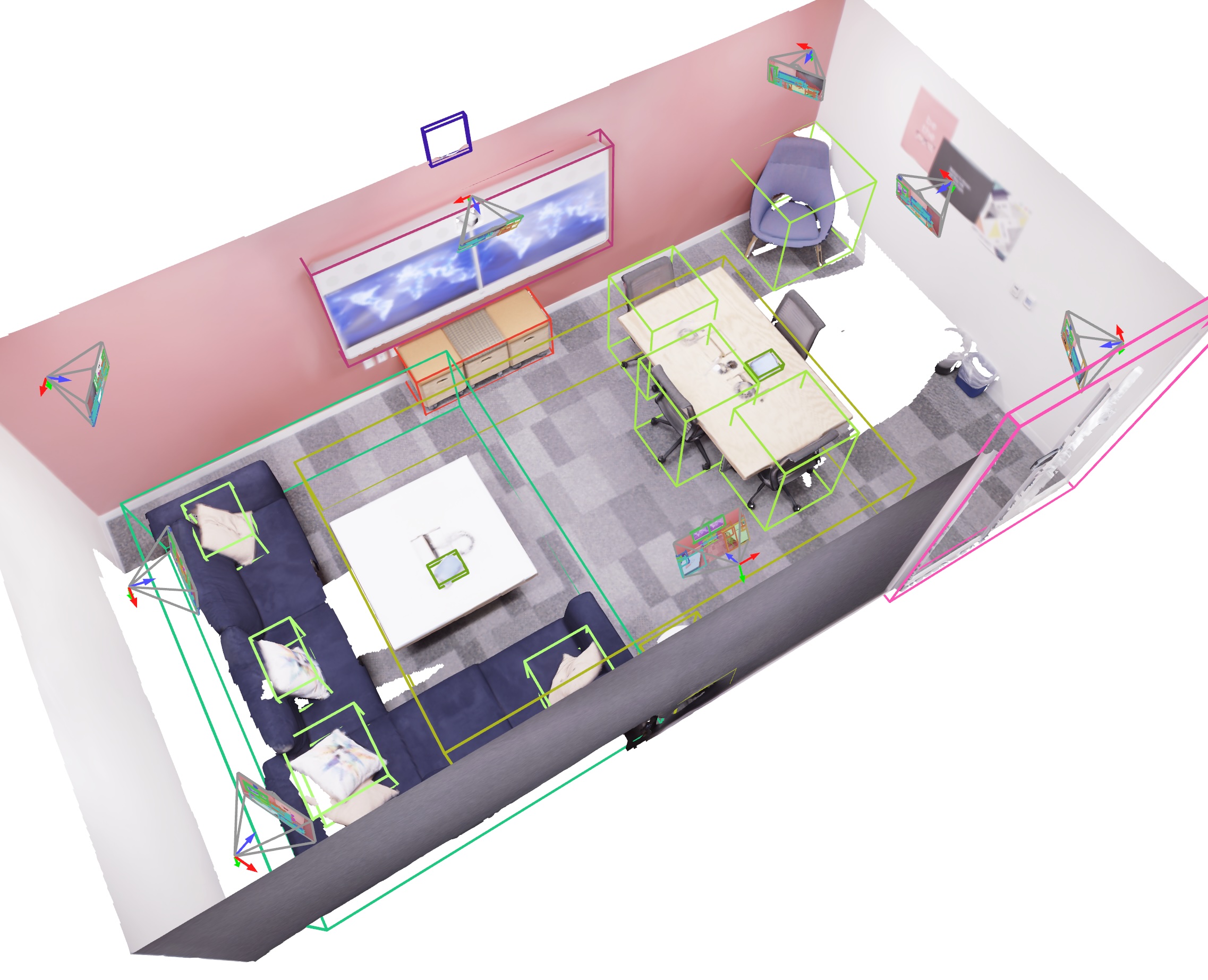}} &
   \raisebox{-0.5\height}{\includegraphics[width=0.33\textwidth]{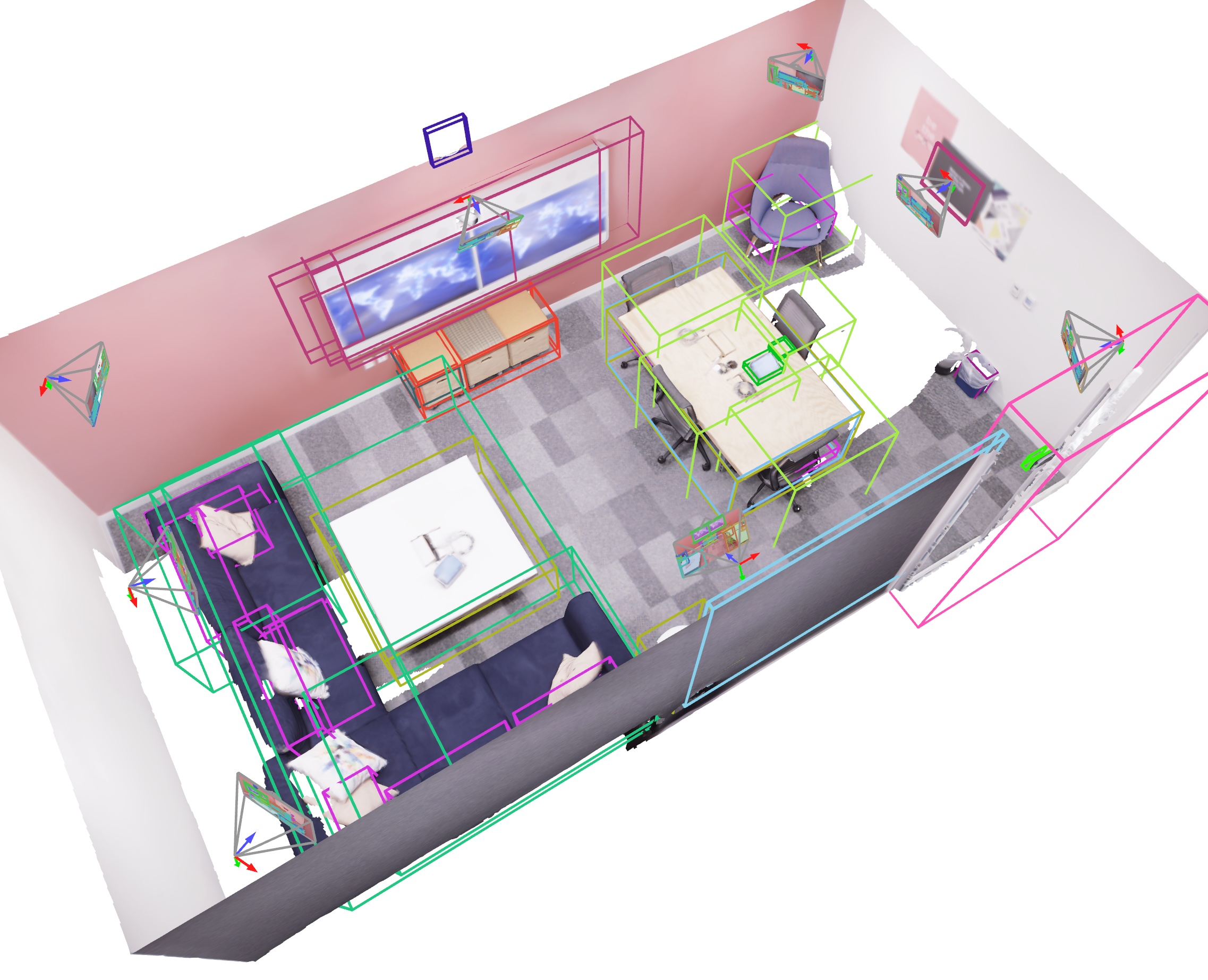}}
  \vspace{1mm}\\
   \rotatebox[origin=c]{90}{} &
   Ground truth &
   OpenIns3D &
   SMOV3D
 \end{tabular}
\caption{\textbf{Qualitative results with few views results on Replica.} Zero-shot 3D object detection, using the Replica categories as prompts.
}
\label{fig:replica}
\vspace{-6pt}
\end{figure*}

%% file: tables/tab_ablation_refinement.tex
\begin{table}[h]
\small
\centering
\begin{tabular}[t]{l|ccc|r}
\toprule[1pt]
Depth refinement & $\mathcal{L}_{depth}$ & $\mathcal{L}_{rgb}$ & $\mathcal{L}_{sim}$            & mAP$_{25}$\T\B\\
\midrule[.5pt]
Global scale & &  & \cmark & 25.7\T\B\\
\midrule[.2pt]
\multirow{4}{*}{Per-mask}& \cmark & & & 24.1\T\\ 
& & \cmark & & 26.1   \\ 
& & & \cmark & 27.8\\
& & \cmark & \cmark & \textbf{28.9}\B\\
\bottomrule[1pt]
\end{tabular}
\caption{\label{table:ablation_refinement}\textbf{Ablation on different aspects of depth refinement.} Evaluated on ScanNet10 (mAP$_{25}$, \%). Per-mask depth refinement using the photometric and CLIP consistency losses yields the best results.
}
\end{table}

%% file: sec/5_conclusion.tex
\section{Limitations and Future Work}
\label{sec:limitations}
Under severe occlusions, some objects are only visible in one view.
While global depth initialization provides some robustness, severe cases can lead to mis-localization during per-mask refinement.

Due to motion blur, occlusions or bad lighting conditions, some objects may be classified differently across views, causing overlapping 3D detections.
Some occurrences of this problem can be seen in \cref{fig:qualitative} and \cref{fig:replica}, e.g. for the "chair" and "couch" classes.
Future work could incorporate more sophisticated fusion logic.

Our method relies on accurate camera poses for back-projection. Its performance may degrade with noisy camera calibrations, a factor we have not explored in this work. 

There remains a performance gap between our RGB-only method and methods using ground-truth depth.
While \OURS narrows this gap significantly, improving monocular depth estimation or developing new refinement strategies to close it further are promising directions for future research.

\section{Conclusion}
\label{sec:conclusion}
In this work, we have conducted an in-depth study on the effectiveness of 2D foundation models for open-vocabulary 3D detection from sparse RGB views.
Our proposed baseline, \OURS, demonstrates that a straightforward, training-free approach can achieve highly competitive results, particularly in challenging sparse-view and long-tail scenarios where trained methods may struggle.
Our work shows that dense 3D geometry is not always a prerequisite for accurate 3D perception, and that
the generalized knowledge embedded in 2D foundation models represents a powerful and practical resource that can be leveraged directly.
Further, by not having any learned 3D component, our method does not rely on having access to 3D data for training or fine-tuning, making it easy to apply in new settings.

{\flushleft \bf Acknowledgment}

This work was partially supported by the Wallenberg AI, Autonomous Systems and Software Program (WASP) funded by the Knut and Alice Wallenberg Foundation.